\def\year{2022}\relax
\documentclass[letterpaper]{article} 
\usepackage[whole]{bxcjkjatype}

\usepackage{aaai22}  
\usepackage{times}  
\usepackage{helvet}  
\usepackage{courier}  
\usepackage[hyphens]{url}  
\usepackage{graphicx} 
\usepackage{natbib}  
\usepackage{caption} 
\DeclareCaptionStyle{ruled}{labelfont=normalfont,labelsep=colon,strut=off} 
\usepackage{algorithm}
\usepackage{algpseudocode}

\usepackage{amsmath,amssymb}
\usepackage{bm}
\usepackage{mathtools}
\usepackage{color}

\usepackage{booktabs}
\usepackage{multirow}

\usepackage{newfloat}
\usepackage{listings}
\floatstyle{ruled}
\newfloat{listing}{tb}{lst}{}
\floatname{listing}{Listing}
\title{Fully Spiking Variational Autoencoder}
\author {
    Hiromichi Kamata \textsuperscript{\rm 1},
    Yusuke Mukuta \textsuperscript{\rm 1,2},
    Tatsuya Harada \textsuperscript{\rm 1,2}
}
\affiliations {
    \textsuperscript{\rm 1} The University of Tokyo\\
    \textsuperscript{\rm 2} RIKEN\\
    \{kamata, mukuta, harada\}@mi.t.u-tokyo.ac.jp
}

\usepackage{bibentry}

\begin{document}

\maketitle

\begin{abstract}
Spiking neural networks (SNNs) can be run on neuromorphic devices with ultra-high speed and ultra-low energy consumption because of their binary and event-driven nature. Therefore, SNNs are expected to have various applications, including as generative models being running on edge devices to create high-quality images. In this study, we build a variational autoencoder (VAE) with SNN to enable image generation. VAE is known for its stability among generative models; recently, its quality advanced. In vanilla VAE, the latent space is represented as a normal distribution, and floating-point calculations are required in sampling. However, this is not possible in SNNs because all features must be binary time series data. Therefore, we constructed the latent space with an autoregressive SNN model, and randomly selected samples from its output to sample the latent variables. This allows the latent variables to follow the Bernoulli process and allows variational learning. Thus, we build the Fully Spiking Variational Autoencoder where all modules are constructed with SNN. To the best of our knowledge, we are the first to build a VAE only with SNN layers. We experimented with several datasets, and confirmed that it can generate images with the same or better quality compared to conventional ANNs. The code is available at \url{https://github.com/kamata1729/FullySpikingVAE}.
\end{abstract}


\bigskip
\section{Introduction}
Recently, artificial neural networks (ANNs) have been evolving rapidly, and have achieved considerable success in computer vision and NLP. However, ANNs often require significant computational resources, which is a challenge in situations where computational resources are limited, such as on edge devices.

Spiking neural networks (SNNs) are neural networks that more accurately mimic the structure of a biological brain than ANNs; notably, SNNs are referred to as the third generation of artificial intelligence \cite{thirdgen}. In a SNN, all information is represented as binary time series data, and is driven by event-based processing. Therefore, SNNs can run with ultra-high speed and ultra-low energy consumption on neuromorphic devices, such as Loihi \cite{loihi}, TrueNorth \cite{truenorth}, and Neurogrid \cite{neurogrid}. For example, on TrueNorth, the computational time is approximately 1/100 lower and the energy consumption is approximately 1/100,000 times lower than on conventional ANNs \cite{Cassidy2014RealTimeSC}.

With the recent breakthroughs on ANNs, research on SNNs has been progressing rapidly. Additionally, SNNs are outperforming ANNs in accuracy in MNIST, CIFAR10, and ImageNet classification tasks \cite{zheng2020going, tsslbp}. Moreover, SNNs are used for object detection \cite{spikingyolo}, sound classification \cite{soundclassification}, optical flow estimation \cite{opticalflow}; however, their applications are still limited.

\begin{figure}[t]
    \centering
    \includegraphics[width=\columnwidth]{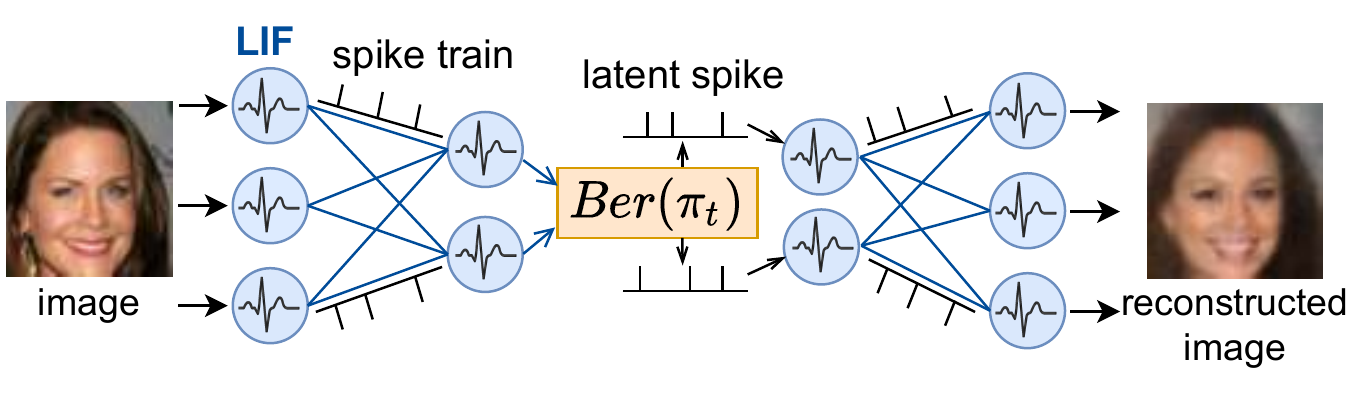} 
    \caption{Illustration of our FSVAE. Entire model is constructed with SNNs. All features are represented as spike trains, and the latent spike trains follow Bernoulli processes.}
    \label{fig:top}
\end{figure}

In particular, image generation models based on SNNs have not been studied sufficiently. Spiking GAN \cite{spikinggan} built a generator and discriminator with shallow SNNs, and generated images of handwritten digits by adversarial learning. However, its generation quality was low, and some undesired images were generated that could not be interpreted as numbers. In \cite{dib}, SNN was used as the encoder and ANN as the decoder to build a VAE \cite{vae}; however, the main focus of their research was efficient spike encoding, and not the image generation task.

In ANNs, image generation models have been extensively studied and can generate high-quality images \cite{vqvae2, stylegan2}. However, in general, image generation models are computationally expensive, and some problems must be solved for edge devices, or for real-time generation. If SNNs can generate images comparably to ANNs, their high speed and low energy consumption can solve these problems.

Therefore, we propose Fully Spiking Variational Autoencoder (FSVAE), which can generate images with the same or better quality than ANN. VAEs are known for their stability among generative models and are related to the learning mechanism of the biological brain \cite{Han214247}. Hence, it is compatible to build VAE with SNN.
In our FSVAE, we built the entire model in SNN, so that it can be implemented in a neuromorphic device in the future. We conducted experiments using MNIST \cite{mnist}, FashionMNIST \cite{fashionmnist}, CIFAR10 \cite{cifar10}, and CelebA \cite{celeba}, and confirmed that \textbf{FSVAE can generate images of equal or better quality than ANN VAE of the same structure}. FSVAE can be implemented on neuromorphic devices in the future, and is expected to improve in terms of speed and energy consumption.

The most difficult aspect of creating VAEs in SNNs is how to create the latent space. In ANN VAEs, the latent space is often represented as a normal distribution. However, within the framework of SNNs, sampling from a normal distribution is not possible because all features must be binary time series data. Therefore, we propose the autoregressive Bernoulli spike sampling. First, we incorporated the idea of VRNN \cite{vrnn} into SNN, and built prior and posterior models with autoregressive SNNs. The latent variables are randomly selected from the output of the autoregressive SNNs, which enables sampling from the Bernoulli processes. This can be realized on neuromorphic devices because it does not require floating-point calculations during sampling as in ANNs, and sampling using a random number generator is possible on actual neuromorphic devices \cite{truenorthrandom, loihi}. In addition, the latent variables can be sampled sequentially; thus, they can be input to the decoder incrementally, which saves time.

The main contributions of this study are summarized as follows.

\begin{itemize}
    \item We propose the autoregressive Bernoulli spike sampling, which uses autoregressive SNNs and constructs the latent space as Bernoulli processes. This sampling method is feasible within the framework of SNN.
    \item We propose Fully Spiking Variational Autoencoder (FSVAE), where all modules are constructed in the SNN.
    \item We experimented with multiple datasets; FSVAE could generate images of equal or better quality than ANN VAE of the same architecture.
\end{itemize}

\section{Related Work}
\subsection{Development of SNNs}

SNNs are neural networks that accurately mimic the structure of the biological brain. In the biological brain, information is transmitted as spike trains (binary time series data with only on/off). This information is transmitted between neurons via synapses, and subsequently, the neuron's membrane potential changes. When it exceeds a threshold, it fires and becomes a spike train to the next neuron.

SNNs mimic these characteristics of the biological brain, modeling biological neurons using differential equations and representing all features as spike trains. This allows SNNs to run faster and asynchronously, because they require fewer floating-point computations and only need computations when the input spike arrives. SNNs can be considered as recurrent neural networks (RNNs) with the membrane potential as its internal state.

Learning algorithms for SNNs have been studied extensively recently. \cite{diehlcook} used a two-layer SNN to recognize MNIST with STDP, an unsupervised learning rule, and achieved 95\% accuracy. Later, \cite{wu2019direct} made it possible to train a deep SNN with backpropagation. Recently, \cite{tsslbp} exceeded the ANN's accuracy for MNIST and CIFAR10 \cite{cifar10} in only 5 timesteps (length of spike trains). \cite{zheng2020going} has even higher accuracy in 2 timesteps for CIFAR10 and ImageNet \cite{imagenet}.

\subsection{Spike Neuron Model}

\begin{figure}[t]
    \centering
    \includegraphics[width=\columnwidth]{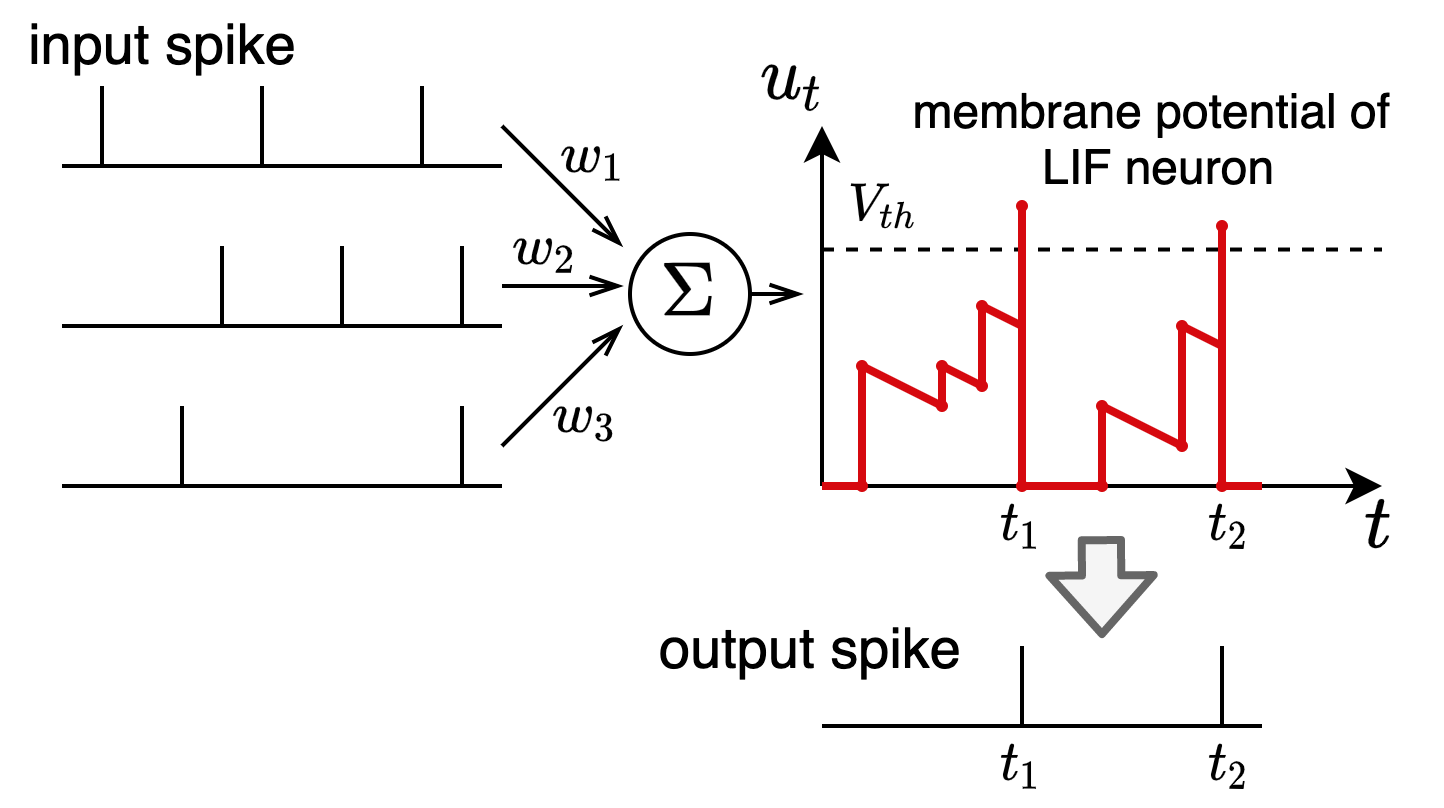} 
    \caption{LIF neuron. When spike trains of the previous layer neurons enter, the internal membrane potential $u_t$ changes by Eq. (\ref{eq:mem}). If $u_t$ exceeds $V_{\mathrm{th}}$, it fires a spike $o_t = 1$; otherwise, $o_t = 0$.}
    \label{fig:lif}
\end{figure}

Although there are several learning algorithms for SNNs, in this study, we follow \cite{zheng2020going}, which currently has the highest recognition accuracy. First, as a neuron model, we use the iterative leaky integrate-and-fire (LIF) model \cite{wu2019direct}, which is a LIF model \cite{lif} solved using the Euler method.
\begin{align}
    u_t = \tau_{\mathrm{decay}} u_{t-1} + x_t
\end{align}
\noindent where $u_t$ is a membrane potential, $x_t$ is a presynaptic input, and $\tau_{\mathrm{decay}}$ is a fixed decay factor.

When $u_t$ exceeds a certain threshold $V_{\mathrm{th}}$, the neuron fires and outputs $o_t=1$. Then, $u_t$ is reset to $u_{rest}=0$. This can be written as follows:
\begin{align}
    u_{t,n} &= \tau_{\mathrm{decay}} u_{t-1,n}(1-o_{t-1,n}) + x_{t,n-1} \label{eq:mem}\\
    o_{t,n} &= H(u_{t,n}-V_{\mathrm{th}}) \label{eq:heaviside}
\end{align}

Here, $u_{t,n}$ is the membrane potential of the $n$th layer, and $o_{t,n}$ is its binary output. $H$ is the heaviside step function. Input $x_{t,n}$ is described as a weighted sum of spikes from neurons in the previous layer,  $x_{t,n-1}=\sum_j w^j o_{t,n-1}^j$. By changing the connection way of $w^j$ , we can implement convolution layers, FC layers, etc.

The next step is to enable learning with backpropagation. As Eq. (\ref{eq:heaviside}) is non-differentiable, we approximate it as follows:
\begin{align}
    \frac{\partial o_{t,n}}{\partial u_{t,n}} = \frac{1}{a} \mathrm{sign}\left(|u_{t,n} - V_{\mathrm{th}}|< \frac{a}{2}\right)
\end{align}

\begin{figure*}[t]
 \begin{center}
  \includegraphics[width=0.9\textwidth]{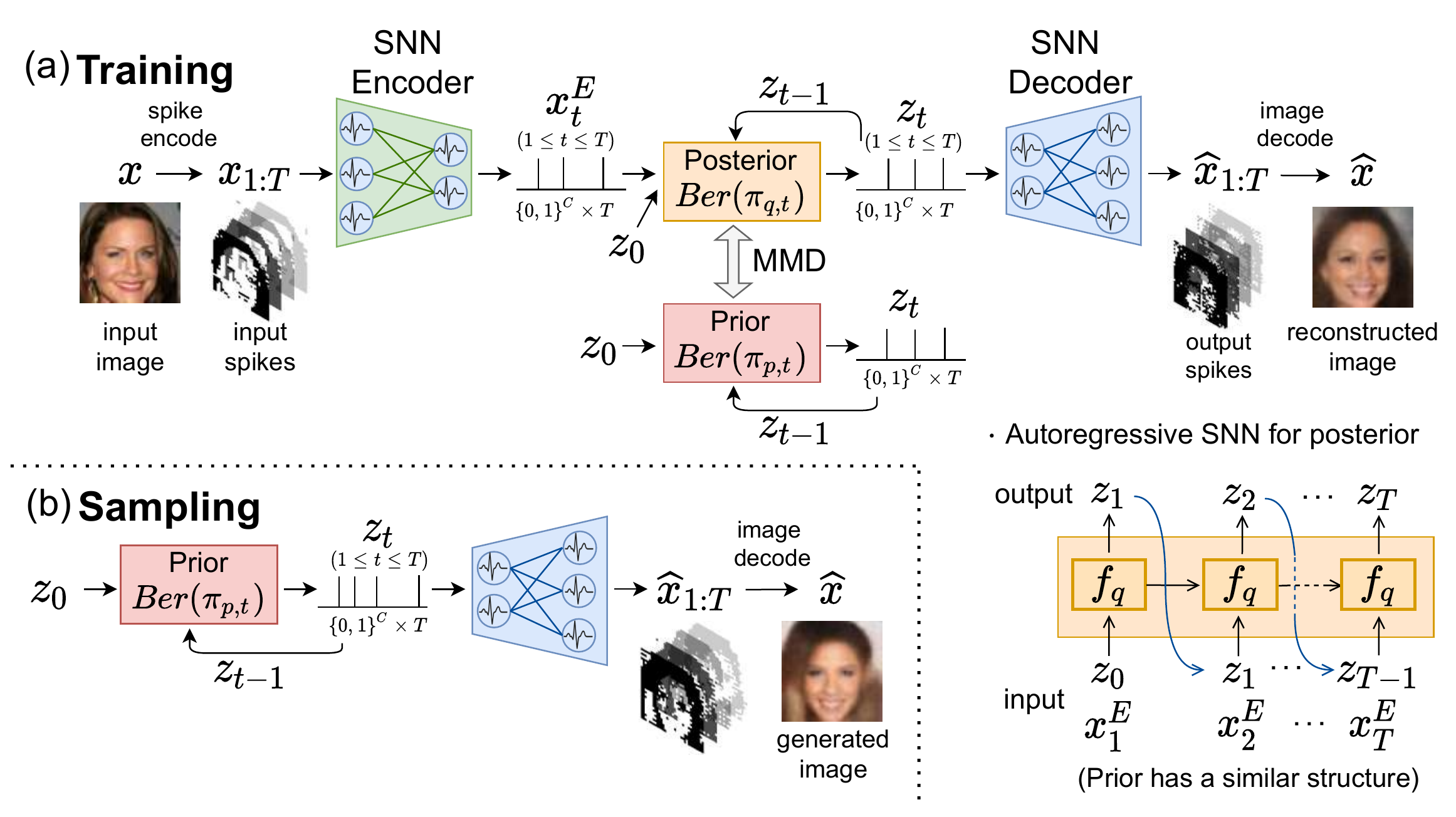}
  \caption{Overview of FSVAE. (a) During the training, the input image $x$ is spike encoded to $\bm{x}_{1:T}$, which is passed through the SNN encoder to obtain $\bm {x}_{1:T}^E$.
  In addition to $\bm{x}_{t}^E$, posterior takes the previously generated latent variables $\bm{z}_{t-1}$ as input, and sequentially outputs $\bm{z}_t$. The lower right figure shows this process in detail. Here, $f_q$ is the SNN model described in Figure \ref{fig:autoreg}.
  In prior, only $\bm{z}_{t-1}$ is used to generate $\bm{z}_t$.
  Next, $\bm{z}_t$ is sequentially input to the SNN decoder, which outputs $\hat{\bm{x}}_{1:T}$ and decodes it to obtain the reconstructed image $\hat{x}$. For the loss, we take the reconstruction error of $x$ and $\hat{x}$ and the MMD of the posterior and prior. (b) During the sampling, the image is generated from $\bm{z}_{1:T}$ sampled in prior.}
  \label{fig:overview}
 \end{center}
\end{figure*}

\subsection{Variational Autoencoder}
Variational Autoencoder (VAE) \cite{vae} is a generative model that explicitly assumes a distribution of the latent variable $z$ over input $x$. Typically, the distribution $p(x|z)$ is represented by deep neural networks, so its inverse transformation is approximated by a simple approximate posterior $q(z|x)$. This allows us to calculate the evidence lower bound (ELBO) of the log likelihood.

\begin{align}
    \log p(x) \geq \mathbb{E}_{q(z|x)}[\log p(x|z)] - \mathrm{KL}[q(z|x)||p(z)]
\end{align}

\noindent where $\mathrm{KL}[Q||P]$ is the Kullback--Leibler (KL) divergence for distributions Q and P. In $q(z|x)$, reparameterization trick is used to sample from $\mathcal{N}(\mu(x),\mathrm{diag}(\sigma(x)^2))$.

VAEs have stable learning among generative models, and can be applied to various tasks, such as anomaly detection \cite{vaeano}. As VAE could generate high-quality images \cite{vqvae2, nvae}, we aimed to build a VAE using SNN.

\subsection{Variational Reccurent Neural Network}
Variational Reccurent Neural Network (VRNN) \cite{vrnn} is a VAE for time series data. Its posterior and prior distributions are set as follows:
\begin{align}
    q(\bm{z}_{1:T} | \bm{x}_{1:T}) &= \prod_{t=1}^T q(\bm{z}_t | \bm{x}_{\leq t}, \bm{z}_{<t}) \\
    p(\bm{z}_{1:T} | \bm{x}_{1:T-1}) &= \prod_{t=1}^T p(\bm{z}_t | \bm{x}_{<t}, \bm{z}_{<t})
\end{align}

Here, $q(\bm{z}_t|\bm{x}_{\leq t},\bm{z}_{<t})$ and $p(\bm{z}_t|\bm{x}_{<t}\bm{z}_{<t})$ are defined using LSTM \cite{lstm}. When sampling, prior inputs $z_{t-1}$ to the decoder to reconstruct $x_{t-1}$, which is used to generate $z_t$ repeatedly.

As SNN is a type of RNNs, we use VRNN to build FSVAE.

\subsection{Generative models in SNN}
Spiking GAN \cite{spikinggan} uses two-layer SNNs to construct a generator and discriminator to train a GAN; however, the quality of the generated image is low. One reason for this is that the time-to-first spike encoding cannot grasp the entire image in the middle of spike trains. In addition, as the learning of SNN is unstable, it would be difficult to perform adversarial learning without regularization.

\subsection{Applying VAE on SNN}
Some studies partially used SNN to create VAEs. In \cite{stewart2021gesture}, human gesture videos captured with a DVS camera were input to an SNN encoder, and the latent variables were generated from the membrane potential of the output neuron; the ANN decoder reconstructed the input from that. Their research aimed to generate pseudo labels for new gestures, not generate images. Moreover, as the decoder was built with ANN, the entire model could not be implemented on neuromorphic devices. Similarly, \cite{dib} had the same problem because their decoder is also built with ANN.

In SVAE \cite{svae}, after training a VAE built on ANN, they converted it to SNN to perform unsupervised learning using STDP, and then, converted it again to ANN to improve the quality of the generated images. Thus, image generation with SNN was not studied.

Some studies used probabilistic neurons to perform variational learning \cite{variationalsnn,probvae}. However, as probabilistic neuron uses the sampling from a Poisson process, it requires additional sampling modules.
Therefore, we created an entire image generation model in deterministic SNN, which is more commonly used.

\subsection{Autoregressive Model for Spike Train Modeling }
When constructing VAEs with SNNs, prior and posterior distributions are required to generate spike trains based on a stochastic process. In a related study, MMD-GLM \cite{mmdglm} modeled a real biological spike train as a Poisson process using an autoregressive ANN. The Mazimum Mean Discrepancy (MMD) was used to measure the consistency with actual spike trains. This is because using KL divergence may cause runaway self-excitation.

In this study, we propose a method for modeling spike trains using an autoregressive SNN in prior and posterior distributions. The details are described in the next section.

\section{Proposed Method}
\subsection{Overview of FSVAE}
The detailed network architecture can be found in \textbf{Supplementary Material A}.

Figure \ref{fig:overview} shows the model overview.
The input image $x$ is transformed into spike trains $\bm{x}_{1:T}$ using direct input encoding \cite{directencode}, and is subsequently input to the SNN encoder. From the output spike trains of Encoder $\bm{x}^E_{1:T}$, the posterior outputs the latent spike trains $\bm{z}_{1:T}$ incrementally. Then, the SNN decoder generates output spike trains $\hat{\bm{x}}_{1:T}$, which is decoded to obtain the reconstructed image $\hat{x}$.
When sampling, prior generates $\bm{z}_{1:T}$ incrementally and inputs it into the SNN decoder to generate image $\hat{x}$.

\subsection{Autoregressive Bernoulli Spike Sampling}

We define the posterior and prior probability distributions of posterior and prior as follows:
\begin{align}
    q(\bm{z}_{1:T} | \bm{x}_{1:T}) &= \prod_{t=1}^T q(\bm{z}_t | \bm{x}_{\leq t}, \bm{z}_{<t}) \\
    p(\bm{z}_{1:T}) &= \prod_{t=1}^T p(\bm{z}_t | \bm{z}_{<t})
\end{align}
We need to model $q(\bm{z}_t | \bm{x}_{\leq t}, \bm{z}_{<t})$ and $p(\bm{z}_t | \bm{z}_{<t})$ with SNNs. Notably, all SNN features must be binary time series data. Therefore, we cannot use the reparameterization trick to sample from the normal distribution as in the conventional VAE.

\begin{figure}[t]
    \centering
    \includegraphics[width=0.95\columnwidth]{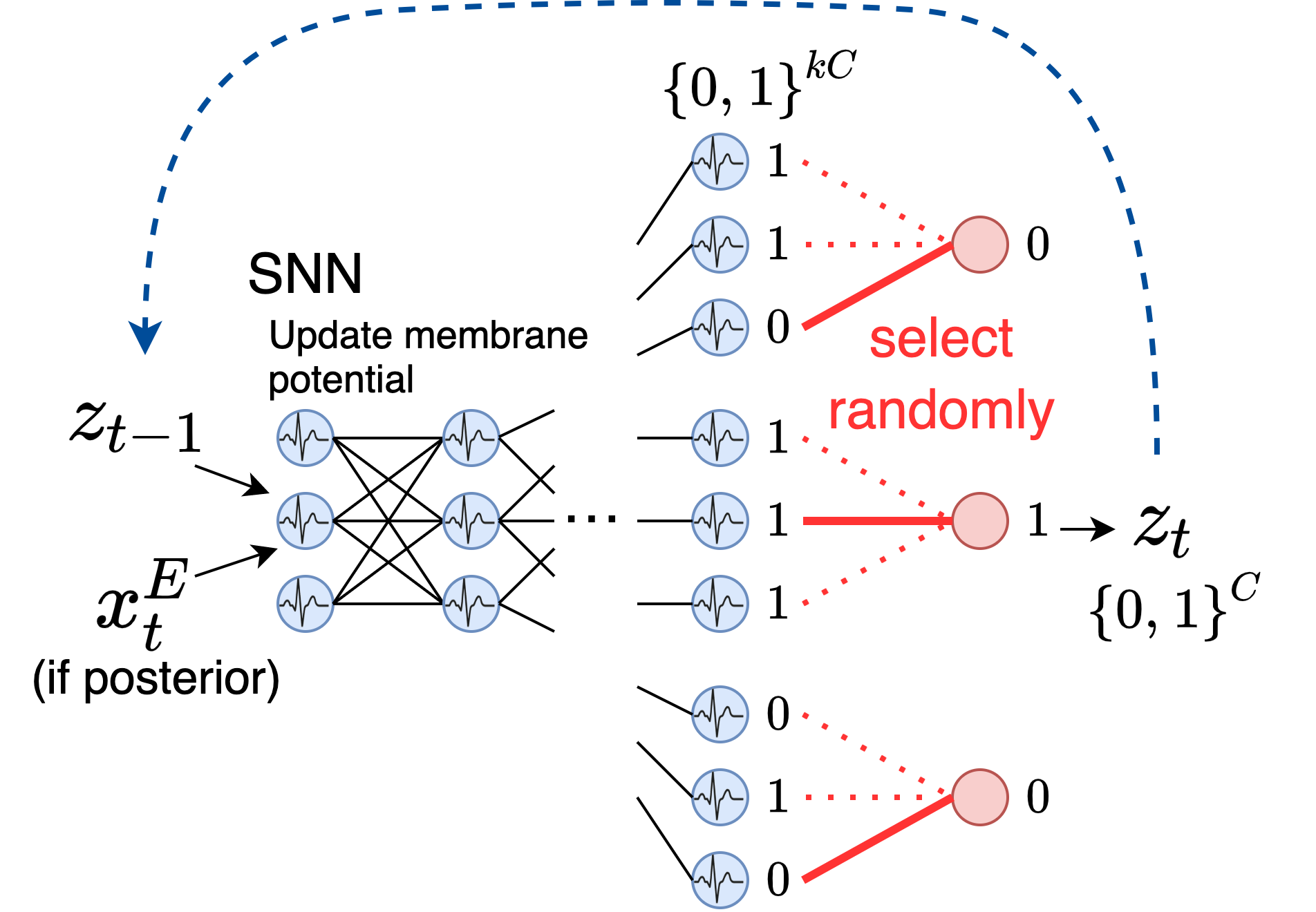} 
    \caption{Autoregressive Bernoulli spike sampling for prior and posterior distributions.
    The input is $\bm{z}_{t-1}$ for prior, and $(\bm{z}_{t-1}, \bm{x}^E_t)$ for posterior. We sample from the Bernoulli distribution by randomly selecting $C$ channels from $kC$ channels of the output.}
    \label{fig:autoreg}
\end{figure}

Consequently, we define $q(\bm{z}_t | \bm{x}_{\leq t}, \bm{z}_{<t})$ and $p(\bm{z}_t | \bm{z}_{<t})$ as Bernoulli distributions which take binary values.
First, we need to generate $\bm{z}_{t}$ sequentially from $\bm{z}_{<t}$; thus, we use the autoregressive SNN model. By randomly selecting one by one from its output, we can sample from a Bernoulli distribution. The overview of this sampling method is shown in Figure \ref{fig:autoreg}.

When the autoregressive SNN model of $q(\bm{z}_t | \bm{x}_{\leq t}, \bm{z}_{<t})$ is $f_q$ and that of $p(\bm{z}_t | \bm{z}_{<t})$ is $f_p$, their outputs are below.
\begin{align}
    \bm{\zeta}_{q,t} \coloneqq & f_q(\bm{z}_{q,t-1}, \bm{x}_t^E; \Theta_{q,t}) \in \{0,1\}^{kC} \\
    \bm{\zeta}_{p,t} \coloneqq & f_p(\bm{z}_{p,t-1}; \Theta_{p,t}) \in \{0,1\}^{kC}
\end{align}

\noindent where $C$ is the dimension of $\bm{z}_t$, $k$ is a natural number starting from 2. $\bm{x}_t^E$ is the output of the encoder. $\Theta_{q,t}$ and $\Theta_{p,t}$ are the sets of membrane potentials of the neurons in $f_q$ and $f_p$, which are updated by the input.

Sampling is performed by randomly selecting one for every $k$ value:

\begin{align}
    \bm{z}_{q,t,c} &= \mathrm{random\_select}(\bm{\zeta}_{q,t}[k(c-1):kc]) \\
    \bm{z}_{p,t,c} &= \mathrm{random\_select}(\bm{\zeta}_{p,t}[k(c-1):kc])
\end{align}

By doing this in $1\leq c\leq C$, we can sample $\bm{z}_{q,t},\bm{z}_{p,t} \in \{0,1\}^{C}$.

In TrueNorth, one of the neuromorphic devices, there is a pseudo random number generator mechanism that can replace the synaptic weight randomly \cite{truenorthrandom}, which makes this sampling method feasible.

This is equivalent to sampling from the following Bernoulli distribution.
\begin{align}
    &\bm{z}_{q,t} | \bm{x}_t, \bm{z}_{q,t-1} \sim Ber(\bm{\pi}_{q,t}) \\
     &\bm{z}_{p,t} | \bm{z}_{p,t-1} \sim Ber(\bm{\pi}_{p,t}) \\
         where&  \left\{
                \begin{array}{ll}
                \bm{\pi}_{q,t,c}=\mathrm{mean}(\bm{\zeta}_{q,t}[k(c-1):kc])\\
                \bm{\pi}_{p,t,c}=\mathrm{mean}(\bm{\zeta}_{p,t}[k(c-1):kc])
                \end{array}
                \right.
\end{align}
Therefore, 
\begin{align}
&q(\bm{z}_t | \bm{x}_{\leq t}, \bm{z}_{<t})=Ber(\bm{\pi}_{q,t}) \\
&p(\bm{z}_t | \bm{z}_{<t})=Ber(\bm{\pi}_{p,t})
\end{align}

\subsection{Spike to Image Decoding using Membrane Potential}

Sampled $\bm{z}_t$ is sequentially input to the SNN decoder, which outputs the spike trains $\hat{\bm{x}}_{1:T} \in \{0,1\}^{C_{\mathrm{out}}\times H\times W\times T}$. We need to decode this into the reconstructed image $\hat{x}\in \mathbb{R}^{C_{\mathrm{out}}\times H\times W}$. To utilize the framework of SNN, we use the membrane potential of the output neuron for spike-to-image decoding. We use non-firing neurons in the output layer, and convert $\hat{\bm{x}}_{1:T}$ into a real value by measuring its membrane potential at the last time $T$. It is given by Eq. (\ref{eq:mem}) as follows:
\begin{align}
    u_T^{\mathrm{out}} &= \tau_{\mathrm{out}} u_{T-1}^{\mathrm{out}} + \hat{x}_T\nonumber \\
        &= \tau_{\mathrm{out}}^2 u_{T-2}^{\mathrm{out}} + \tau_{\mathrm{out}} \hat{x}_{T-1} + \hat{x}_T = \sum_{t=1}^{T} \tau_{\mathrm{out}}^{T-t} \hat{x}_t \label{eq:stidecode}
\end{align}

We set $\tau_{\mathrm{out}}=0.8$, and obtain the real-valued reconstructed image as $\hat{x}=\tanh (u_T^{\mathrm{out}})$.

\subsection{Loss Function}

The ELBO is as follows:
\begin{align}
    ELBO =& \mathbb{E}_{q(\bm{z}_{1:T}|\bm{x}_{1:T})}[\log p(\bm{x}_{1:T}|\bm{z}_{1:T})]\nonumber \\
        & - \mathrm{KL}[q(\bm{z}_{1:T}|\bm{x}_{1:T})||p(\bm{z}_{1:T})] \label{ELBO}
\end{align}
The first term is the reconstruction loss, which is $\mathrm{MSE}(x,\hat{x})$ as in the usual VAE. The second term, KL divergence, represents the closeness of the prior and posterior probability distributions. Traditional VAEs use KL divergence, but we use MMD, which has been shown to be a more suitable distance metric for spike trains in MMD-GLM \cite{mmdglm}. MMD can be written using the kernel function $k$ as follows:

\begin{align}
    &\mathrm{MMD}^2[q(\bm{z}_{1:T}|\bm{x}_{1:T}),p(\bm{z}_{1:T})] \nonumber\\
    =&\underset{\bm{z},\bm{z}'\sim q}{\mathbb{E}}[k(\bm{z}_{1:T},\bm{z}'_{1:T})] +\underset{\bm{z},\bm{z}'\sim p}{\mathbb{E}}[k(\bm{z}_{1:T},\bm{z}'_{1:T})]\nonumber\\
    &-2\underset{\bm{z}\sim q,\bm{z}'\sim p}{\mathbb{E}}[k(\bm{z}_{1:T},\bm{z}'_{1:T})] \label{eq:mmd}
\end{align}

We set $k(\bm{z}_{1:T},\bm{z}'_{1:T})=\sum_t \mathrm{PSP}(\bm{z}_{\leq t})\mathrm{PSP}(\bm{z}'_{\leq t})$, as the model based kernel in MMD-GLM. The $\mathrm{PSP}$ stands for a postsynaptic potential function that can capture the time-series nature of spike trains \cite{superspike}. We use the first-order synaptic model \cite{tsslbp} as the $\mathrm{PSP}$. The following update formula is used to calculate the $\mathrm{PSP}(\bm{z}_{\leq t})$.
\begin{align}
    \mathrm{PSP}(\bm{z}_{\leq t}) = \left( 1-\frac{1}{\tau_{\mathrm{syn}}}\right) \mathrm{PSP}(\bm{z}_{\leq t-1}) + \frac{1}{\tau_{\mathrm{syn}}}\bm{z}_t \label{eq:psp}
\end{align}
where $\tau_{\mathrm{syn}}$ is the synaptic time constant. We set $\mathrm{PSP}(\bm{z}_{\leq 0})=0$. 

This gives us Eq. (\ref{eq:mmd}), as follows. The detailed derivation can be found in \textbf{Supplementary Material B}.

\begin{align}
    &\mathrm{MMD}^2[q(\bm{z}_{1:T}|\bm{x}_{1:T}),p(\bm{z}_{1:T})]\nonumber\\
    =& \sum_{t=1}^T \| \mathrm{PSP}(\underset{\bm{z}\sim q}{\mathbb{E}}[\bm{z}_{\leq t}]) - \mathrm{PSP}(\underset{\bm{z}\sim p}{\mathbb{E}}[\bm{z}_{\leq t}]) \|^2 \\
    =& \sum_{t=1}^T \| \mathrm{PSP}(\bm{\pi}_{q,\leq t}) - \mathrm{PSP}(\bm{\pi}_{p,\leq t})\|^2
\end{align}

The loss function is calculated as follows:
\begin{align}
    \mathcal{L} = \mathrm{MSE}(x,\hat{x}) + \sum_{t=1}^T \| \mathrm{PSP}(\bm{\pi}_{q,\leq t}) - \mathrm{PSP}(\bm{\pi}_{p,\leq t})\|^2 \label{eq:lossmmd}
\end{align}

\begin{table*}[t]
\centering
\begin{tabular}{@{}cccccc@{}}
\toprule
\multirow{2}{*}{Dataset} &
  \multirow{2}{*}{Model} &
  \multirow{2}{*}{\begin{tabular}[c]{@{}c@{}}Reconstruction\\ Loss$\searrow$\end{tabular}} &
  \multirow{2}{*}{\begin{tabular}[c]{@{}c@{}}Inception\\ Score$\nearrow$\end{tabular}} &
  \multicolumn{2}{c}{Fr\'echet Distance$\searrow$} \\ \cmidrule(l){5-6} 
                                                                         &                       &                &                & Inception (FID) & Autoencoder    \\ \midrule
\multirow{2}{*}{MNIST}                                                   & ANN                   & 0.048          & 5.947          & 112.5           & \textbf{17.09} \\
                                                                         & \textbf{FSVAE (Ours)} & \textbf{0.031} & \textbf{6.209} & \textbf{97.06}  & 35.54          \\ \midrule
\multirow{2}{*}{\begin{tabular}[c]{@{}c@{}}Fashion\\ MNIST\end{tabular}} & ANN                   & 0.050          & 4.252          & 123.7           & 18.08          \\
                                                                         & \textbf{FSVAE (Ours)} & \textbf{0.031} & \textbf{4.551} & \textbf{90.12}  & \textbf{15.75} \\ \midrule
\multirow{2}{*}{CIFAR10}                                                 & ANN                   & 0.105          & 2.591          & 229.6           & 196.9          \\
                                                                         & \textbf{FSVAE (Ours)} & \textbf{0.066} & \textbf{2.945} & \textbf{175.5}  & \textbf{133.9} \\ \midrule
\multirow{2}{*}{CelebA}                                                  & ANN                   & 0.059          & 3.231          & \textbf{92.53}  & 156.9          \\
                                                                         & \textbf{FSVAE (Ours)} & \textbf{0.051} & \textbf{3.697} & 101.6           & \textbf{112.9} \\ \bottomrule
\end{tabular}
\caption{Results for each dataset. In all datasets, our model outperforms ANN in the inception score. Moreover, our model outperforms MNIST and Fashion MNIST in FID, CIFAR10 in all metrics, and CelebA in Autoencoder's Fr\'echet distance. Reconstruction losses are better for our model in all datasets.}
\label{tab:results}
\end{table*}

\section{Experiments}
We implemented FSVAE in PyTorch \cite{pytorch}, and evaluated it using MNIST, Fashion MNIST, CIFAR10, and CelebA. The results are summarized in Table \ref{tab:results}.

\subsection{Datasets}
For MNIST and Fashion MNIST, we used 60,000 images for training and 10,000 images for evaluation. The input images were resized to 32$\times$32. For CIFAR10, we used 50,000 images for training and 10,000 images for evaluation. For CelebA, we used 162,770 images for training and 19,962 images for evaluation. The input images were resized to 64$\times$64.

\subsection{Network Architecture}

The SNN encoder comprises several convolutional layers, each with kernel\_size=3 and stride=2. The number of layers is 4 for MNIST, Fashion MNIST and CIFAR10, and 5 for CelebA. After each layer, we set a tdBN \cite{zheng2020going}, and then, input the feature to the LIF neuron to obtain the output spike trains. The encoder's output is  $\bm{x}_t^E\in\{0,1\}^C$ with latent dimension $C=128$. We combine it with $\bm{z}_{t-1}\in\{0,1\}^C$, and input it to the posterior model; thus, its input dimension is $128+128=256$. Additionally, we set $\bm{z}_{0}=\bm{0}$. The posterior model comprises three FC layers, and increases the number of channels by a factor of $k=20$. $\bm{z}_t$ is sampled from it by the proposed autoregressive Bernoulli spike sampling. We repeat this process $T=16$ times. The prior model has the same model architecture; however, as the input is only $\bm{z}_{t-1}$, its input dimension is 128.

Sampled $\bm{z}_t$ is input to the SNN decoder, which contains the same number of deconvolution layers as the encoder. Decoder's output is spike trains of the same size as the input. Finally, we perform spike-to-image decoding to obtain the reconstructed image, according to Eq. (\ref{eq:stidecode}).

\subsection{Training Settings}
We use AdamW optimizer \cite{adamw}, which trains 150 epochs with a learning rate of 0.001 and a weight decay of 0.001. The batch size is 250. In prior model, teacher forcing \cite{teacherforcing} is used to stabilize training, so that the prior's input is $\bm{z}_{q,t}$, which is sampled from the posterior model. In addition, to prevent posterior collapse, scheduled sampling \cite{scheduled} is performed. With a certain probability, we input $\bm{z}_{p,t}$ to the prior instead of $\bm{z}_{q,t}$. This probability varies linearly from 0 to 0.3 during training.

\begin{figure*}[t]
 \begin{center}
  \includegraphics[width=0.95\textwidth]{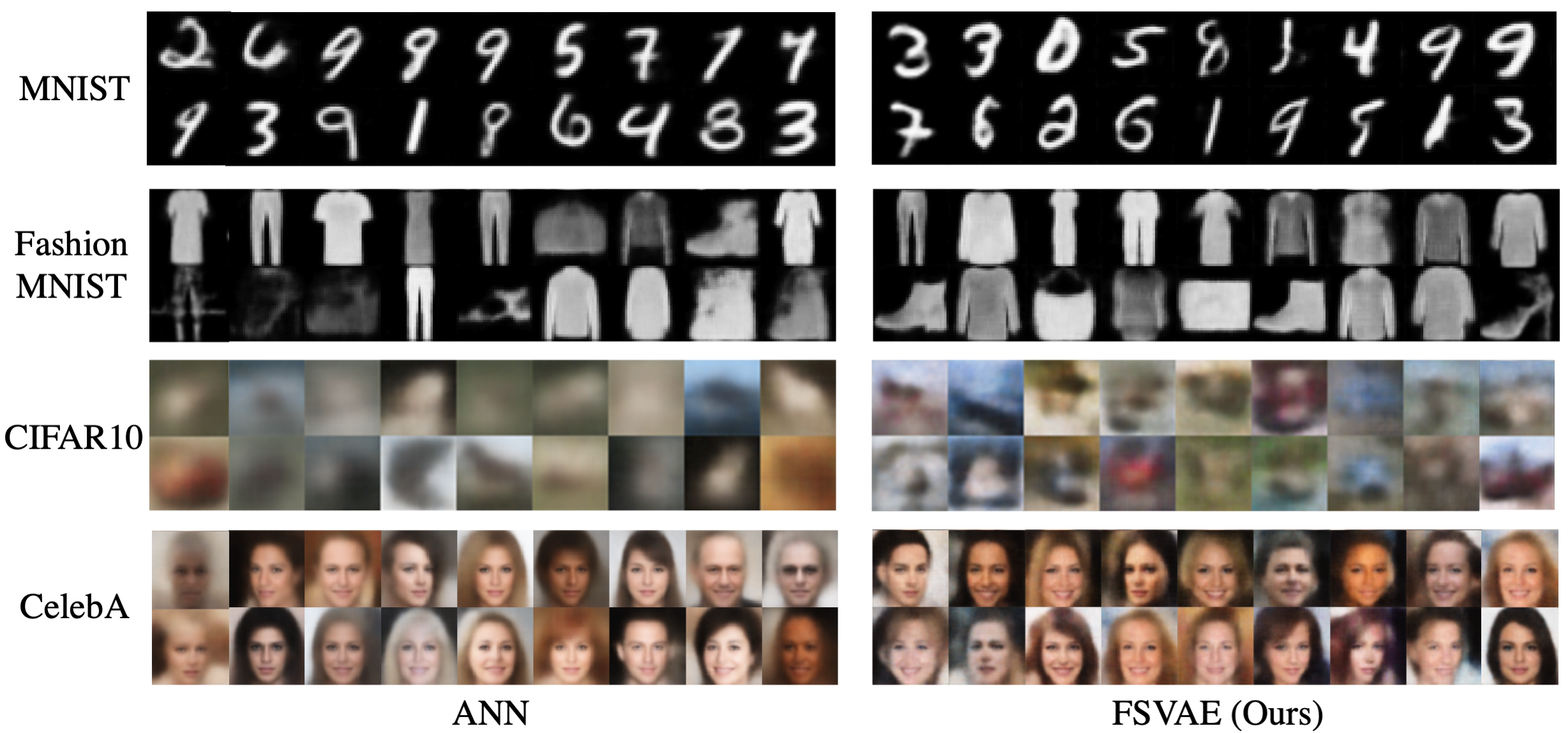}
  \caption{Generated images of ANN VAE and our FSVAE (SNN).}
  \label{fig:samples}
 \end{center}
\end{figure*}

\subsection{Evaluation Metrics}
The quality of the sampled images is measured by the inception score \cite{inceptionscore} and FID \cite{fid, cleanfid}. However, as FID considers the output of the ImageNet pretrained inception model, it may not work well on datasets such as MNIST, which have a different domain from ImageNet. Consequently, we trained an autoencoder on each dataset beforehand, and used it to measure the Fr\'echet distance of the autoencoder's latent variables between sampled and real images. We sampled 5,000 images to measure the distance. As a comparison method, we prepared vanilla VAEs of the same network architecture built with ANN, and trained on the same settings.

Table \ref{tab:results} shows that our FSVAE outperforms ANN in the inception score for all datasets. For MNIST and Fashion MNIST, our model outperforms in terms of FID, for CIFAR10 in all metrics, and for CelebA in the Fr\'echet distance of the pretrained autoencoder. As SNNs can only use binary values, it is difficult to perform complex tasks; in contrast, our FSVAE achieves equal or higher scores than ANNs in image generation.

\subsubsection{Computational Cost}
We measured how many floating-point additions and multiplications were required for inferring a single image. We summarized the results in Table \ref{tab:complex}. The number of additions is 6.8 times less for ANN, but the number of multiplications is 14.8 times less for SNN. In general, multiplication is more expensive than addition, so fewer multiplications are preferable. Moreover, as SNNs are expected to be about 100 times faster than ANNs when implemented in neuromorphic device, FSVAE can significantly outperform ANN VAE in terms of speed.

\begin{table}[H]
\centering
\begin{tabular}{@{}ccc@{}}
\toprule
\multirow{2}{*}{Model} & \multicolumn{2}{c}{Computational complexity} \\ \cmidrule(l){2-3} 
                       & Addition               & Multiplication      \\ \midrule
ANN                    & $\bm{7.4\times 10^9}$       & $7.4\times 10^9$    \\
FSVAE (Ours)              & $5.0\times 10^{10}$    & $\bm{5.6\times 10^8}$    \\ \bottomrule
\end{tabular}
\caption{Comparison of the amount of computation required to infer a single image in MNIST.}
\label{tab:complex}
\end{table}

\begin{figure}[h]
    \centering
    \includegraphics[width=0.9\columnwidth]{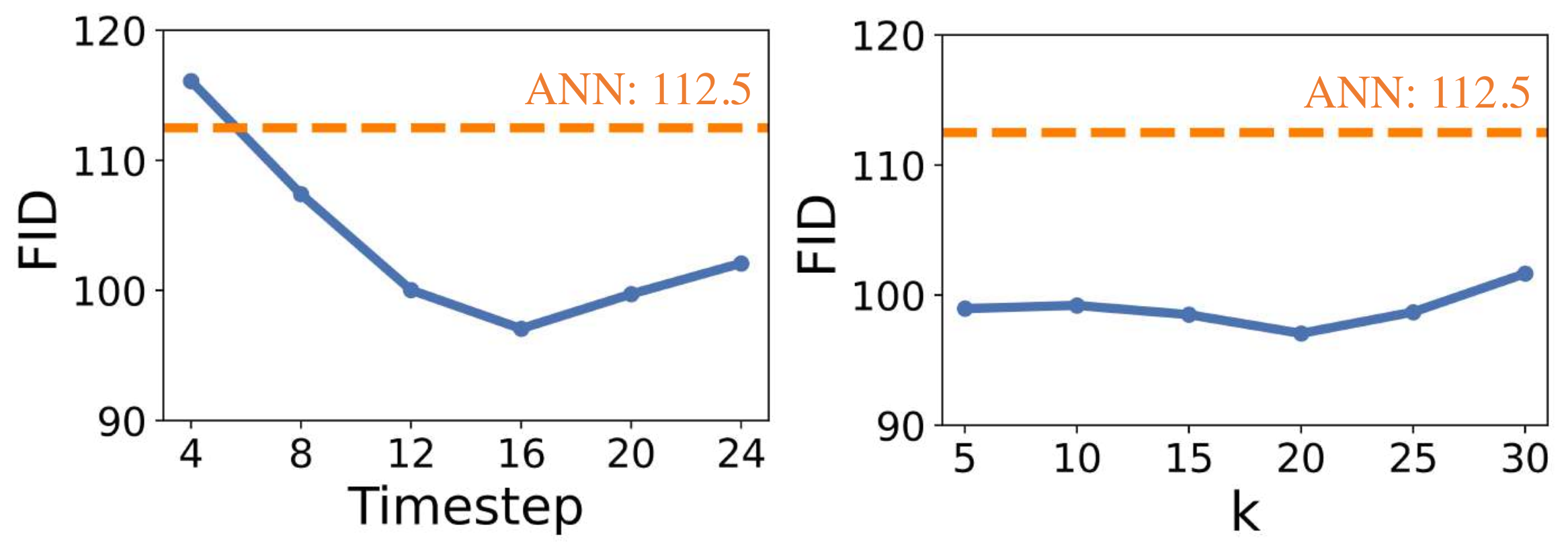} 
    \caption{Left: relationship between the timestep and FID score of generated images in MNIST. Right: relationship between $k$ (multiplier for the channel) and FID in MNIST.}
    \label{fig:tkfid}
\end{figure}

Figure \ref{fig:tkfid} shows the change in FID depending on the timestep and $k$, the number of output choices. The best FID was obtained when the timestep was 16. A small number of timesteps does not have enough expressive power, whereas a large number of timesteps makes the latent space too large; thus, the best timestep was determined to be 16. Moreover, FID does not change so much with $k$, but it is best when $k=20$.

\subsubsection{Ablation Study} The results are summarized in Table \ref{tab:ablation}. When calculating KLD, $\epsilon=0.01$ was added to $\bm{\pi}_{q,t}$ and $\bm{\pi}_{p,t}$ to avoid divergence. The best FID score was obtained using MMD as the loss function and applying $\mathrm{PSP}$ to its kernel. 

\begin{table}[h]
\centering
\begin{tabular}{@{}ccc|c@{}}
\toprule
KLD        & MMD        & apply $\mathrm{PSP}$  & FID$\searrow$             \\ \midrule
\checkmark &            &            & 114.3                     \\
           & \checkmark &            & 106.0                     \\
           & \checkmark & \checkmark & \textbf{101.6} \\ \bottomrule
\end{tabular}
\caption{Ablation study on CelebA. KLD and MMD indicate the distance metric used in the loss function. We also measured whether to apply $\mathrm{PSP}$ to the MMD kernel.}
\label{tab:ablation}
\end{table}

\subsection{Qualitative Evaluation}
Figure \ref{fig:samples} shows examples of the generated images. In CIFAR10 and Fashion MNIST, the ANN VAE generated blurry and hazy images, whereas our FSVAE generated clearer images. In CelebA, FSVAE generated images with more distinct background areas than ANN. This is because the latent variables of FSVAE are discrete spike trains, thus it can avoid posterior collapse, as VQ-VAE \cite{vqvae}.

\begin{figure}[t]
    \centering
    \includegraphics[width=\columnwidth]{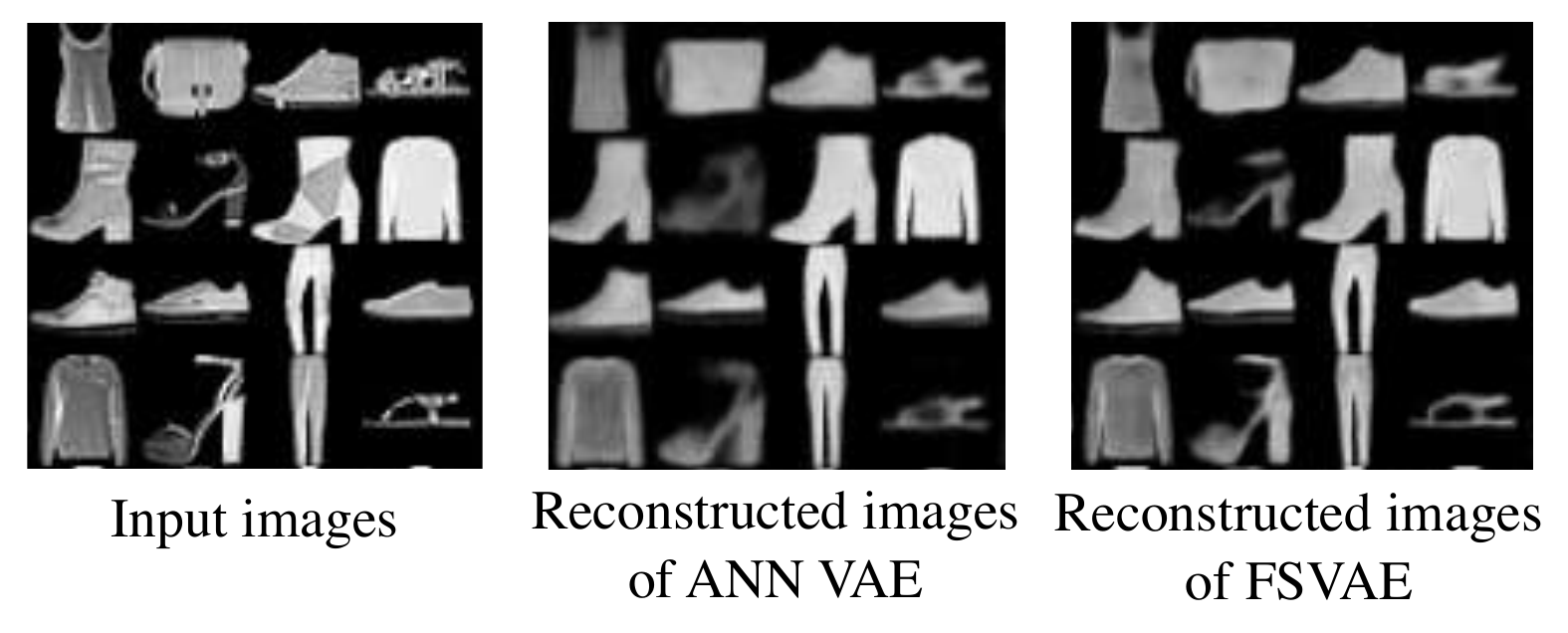} 
    \caption{Reconstructed images of Fashion MNIST.}
    \label{fig:recons}
\end{figure}

Figure \ref{fig:recons} shows the reconstructed images of Fashion MNIST. FSVAE reconstruct images more clearly than ANN, especially the details. This is because that the latent variable of FSVAE is discrete. Posterior collapse is caused by the latent variables being ignored by the decoder. With discrete latent variables, every latent variable becomes meaningful, which can prevent posterior collapse.

\section{Conclusion}
In this study, we proposed FSVAE, which allows image generation on SNNs with equal or higher quality than ANNs. We proposed autoregressive Bernoulli spike sampling, which is a spike sampling strategy that can be implemented on neuromorphic devices. This sampling method is used in prior and posterior distribution, and we model the latent spike trains as a Bernoulli process. Experiments on multiple datasets show that FSVAE can generate images with equal or higher quality than ANN VAE of the same architecture. As SNNs can be significantly faster on neuromorphic devices, FSVAE can significantly outperform ANN VAE in terms of speed. In the future, when we incorporate the recent VAE researches, we will soon be able to achieve high-resolution image generation with SNNs.

\section{Acknowledgments}
This work was partially supported by JST AIP Acceleration Research JPMJCR20U3, Moonshot R\&D Grant Number JPMJPS2011, CREST Grant Number JPMJCR2015, and Basic Research Grant (Super AI) of Institute for AI and Beyond of the University of Tokyo. We would like to thank Yang Li and Muyuan Xu for some helpful advice of SNNs. We also thank Tomoyuki Takahata, Yusuke Mori, and Atsuhiro Noguchi for helpful discussions.

\bibliography{aaai22.bib}

\begin{figure*}[t!]
 \centering
  \includegraphics[width=0.9\textwidth]{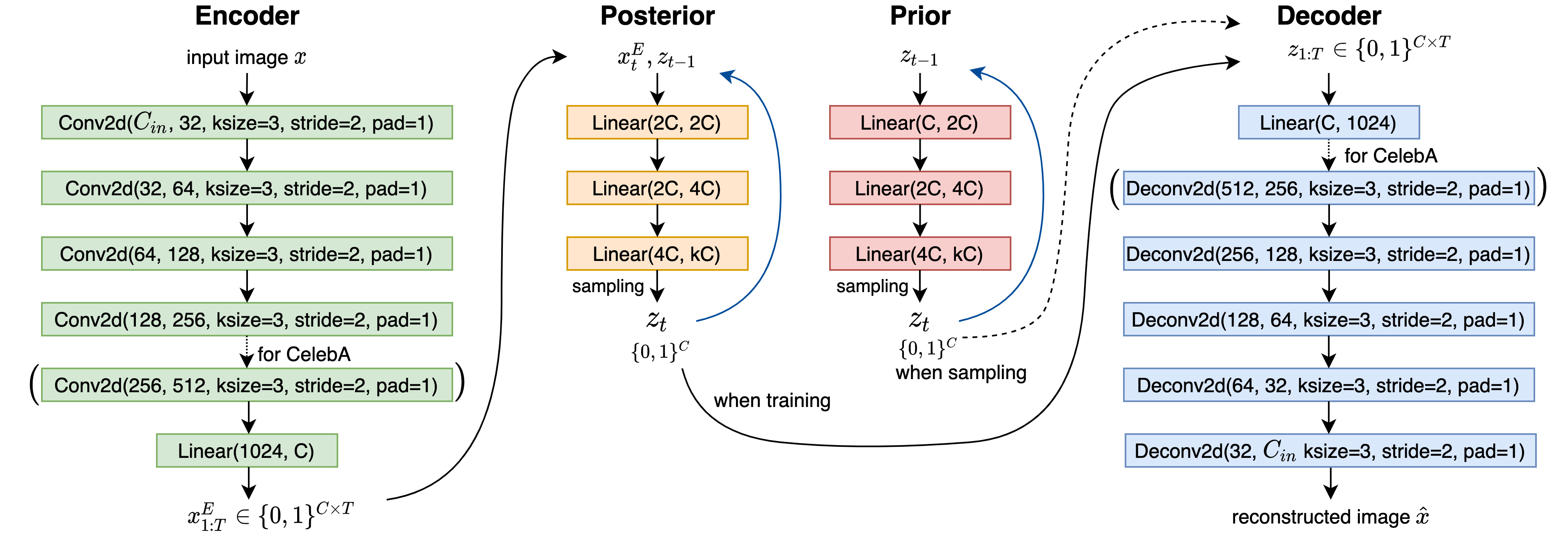}
  \caption{Detailed architecture of FSVAE. After each layer a tdBN layer LIF neuron is inserted. In our settings, timstep $T$ is set to 16, latent dimension $C$ to 128, and $k$ to 20.}
  \label{fig:architecture}
\end{figure*}
\section{Supplementary Material}
\section{A\quad Network Architecture}

Figure \ref{fig:architecture} shows the detailed architecture of our FSVAE. For the CelebA dataset we added an additional layer to both encoder and decoder. The architecture of ANN VAE is the same as FSVAE, except for prior and posterior.

\section{B\quad Derivation of MMD}
In this section, we describe the derivation of Eq. (24).

We have $k(\bm{z}_{1:T},\bm{z}'_{1:T})=\sum_t h(\bm{z}_{\leq t})h(\bm{z}'_{\leq t})$, so Eq. (21) is as follows:

\begin{align*}
    &\mathrm{MMD}^2[q(\bm{z}_{1:T}|\bm{x}_{1:T}),p(\bm{z}_{1:T})] \nonumber\\
    =&\sum_t \left(\underset{\bm{z},\bm{z}'\sim q}{\mathbb{E}}[\mathrm{PSP}(\bm{z}_{\leq t})\mathrm{PSP}(\bm{z}'_{\leq t})] \right. \nonumber \\
     &\qquad +\underset{\bm{z},\bm{z}'\sim p}{\mathbb{E}}[\mathrm{PSP}(\bm{z}_{\leq t})\mathrm{PSP}(\bm{z}'_{\leq t})]  \nonumber\\
    &\qquad \left. -2\underset{\bm{z}\sim q,\bm{z}'\sim p}{\mathbb{E}}[\mathrm{PSP}(\bm{z}_{\leq t})\mathrm{PSP}(\bm{z}'_{\leq t})] \right) \nonumber \\
    =&\sum_t \left(\underset{\bm{z}\sim q}{\mathbb{E}}[\mathrm{PSP}(\bm{z}_{\leq t})]^2 + \underset{\bm{z}\sim p}{\mathbb{E}}[\mathrm{PSP}(\bm{z}_{\leq t})]^2 \right.\nonumber\\
     &\qquad \left. -2\underset{\bm{z}\sim q}{\mathbb{E}}[\mathrm{PSP}(\bm{z}_{\leq t})]\underset{\bm{z}\sim p}{\mathbb{E}}[\mathrm{PSP}(\bm{z}_{\leq t})]^2 \right) \nonumber\\
    =& \sum_t \left\| \underset{\bm{z}\sim q}{\mathbb{E}}[\mathrm{PSP}(\bm{z}_{\leq t})] - \underset{\bm{z}\sim p}{\mathbb{E}}[\mathrm{PSP}(\bm{z}_{\leq t})] \right\|^2 \nonumber
\end{align*}

Here, according to Eq. (22), we can obtain $\mathrm{PSP}$ as 
\begin{align*}
    \mathrm{PSP}(\bm{z}_{\leq t}) = \sum_{i=0}^t \frac{1}{\tau_{\mathrm{syn}}}\left(1-\frac{1}{\tau_{\mathrm{syn}}} \right)^i \bm{z}_{t-i}.
\end{align*}

Thus, we have $\mathbb{E}[\mathrm{PSP}(\bm{z}_{\leq t})] = \mathrm{PSP}(\mathbb{E}[\bm{z}_{\leq t}])$. Therefore, MMD is as follows:
\begin{align*}
&\mathrm{MMD}^2[q(\bm{z}_{1:T}|\bm{x}_{1:T}),p(\bm{z}_{1:T})]\nonumber\\
    =& \sum_{t} \left\| \mathrm{PSP}(\underset{\bm{z}\sim q}{\mathbb{E}}[\bm{z}_{\leq t}]) - \mathrm{PSP}(\underset{\bm{z}\sim p}{\mathbb{E}}[\bm{z}_{\leq t}]) \right\|^2 \nonumber\\
    =& \sum_{t} \left\| \mathrm{PSP}(\bm{\pi}_{q,\leq t}) - \mathrm{PSP}(\bm{\pi}_{p,\leq t})\right\|^2\nonumber
\end{align*}

\section{C\quad Derivation of KLD}

In the ablation study, we used KLD as the distance metric between the prior and posterior distribution. The KLD is calculated as follows:

\begin{align*}
    &\mathrm{KL} [q(\bm{z}_{1:T})\| p(\bm{z}_{1:T})] \\
    =& \sum_{t=1}^T \mathrm{KL} [ q(\bm{z}_t | \bm{x}_{\leq t}, \bm{z}_{<t}) \| p(\bm{z}_t | \bm{z}_{<t})]\\
    =& \sum_{t=1}^T \mathrm{KL} [Ber(\bm{\pi}_{q,t} \| Ber(\bm{\pi}_{p,t})] \\
    =& \sum_{t=1}^T \left(\pi_{q,t,c} \log \frac{\pi_{q,t,c}}{\pi_{p,t,c}} + (1-\pi_{q,t,c}) \log \frac{1-\pi_{q,t,c}}{1-\pi_{p,t,c}}\right)
\end{align*}

To avoid divergence, we added $\epsilon = 0.01$ to $\pi_{q,t,c}$ and $\pi_{p,t,c}$.
\section{D\quad Analysis of Spike Activities}
\begin{figure}[htbp]
    \centering
    \includegraphics[width=0.9\columnwidth]{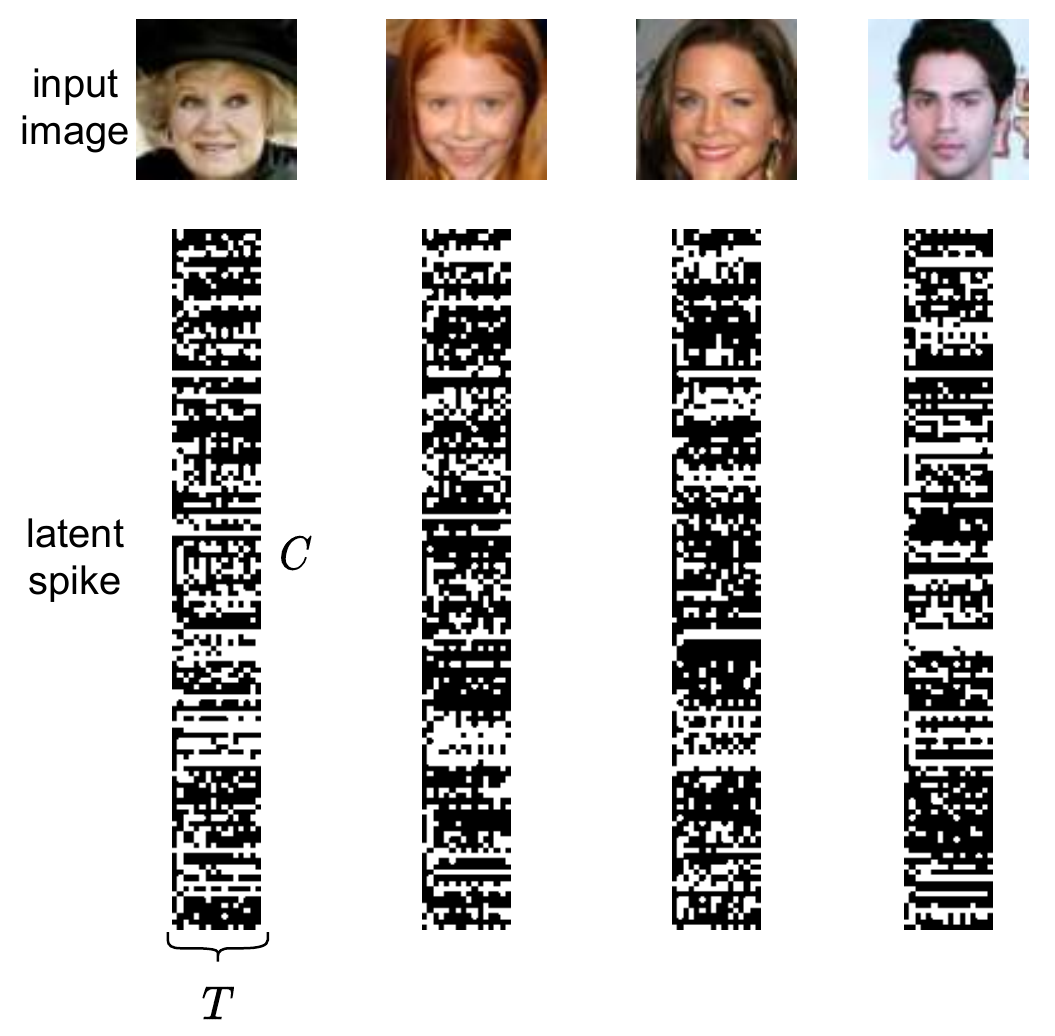} 
    \caption{Input images and their corresponding latent spike trains.}
    \label{fig:latent_and_images}
\end{figure}

Figure \ref{fig:latent_and_images} shows the input images and their corresponding latent spike trains. It can be seen that some spike train fire more periodically while others fire more burst-like. This diversity allows the spike train to encode a variety of different images.

\begin{figure}[htbp]
    \centering
    \includegraphics[width=0.9\columnwidth]{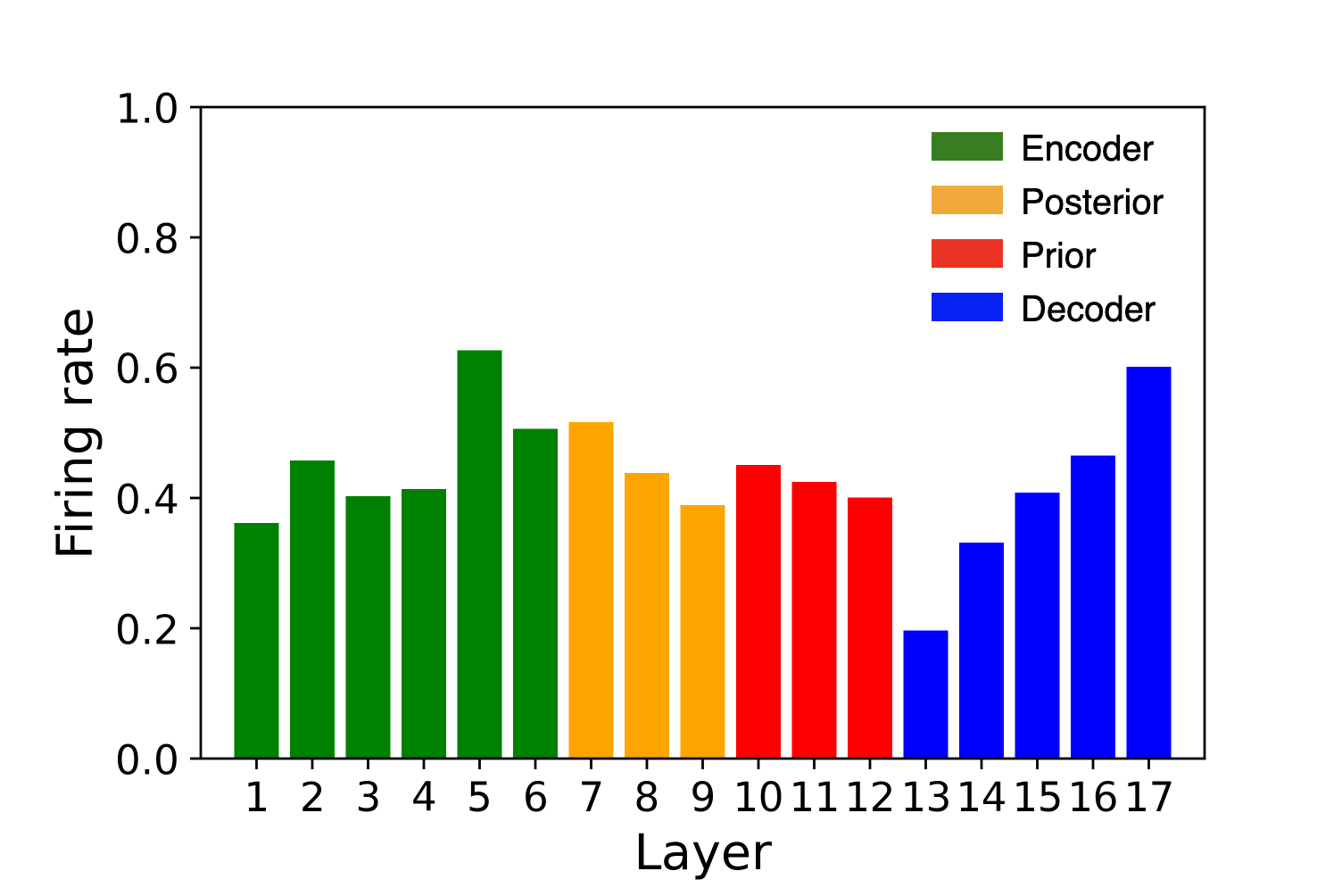} 
    \caption{Firing rates in each layer of FSVAE on CelebA}
    \label{fig:fire_rate}
\end{figure}

Figure \ref{fig:fire_rate} shows the average firing rates of each layer. Each firing rate is stable, and contributes to stable learning. 

\clearpage

\section{E\quad Codes for Algorithms}
We show the detailed code for autoregressive Bernoulli spike sampling for posterior in Algorithm \ref{alg:algorithm1}, and prior in Algorithm \ref{alg:algorithm2}. Additionally, Algorithm \ref{alg:algorithm3} shows the overall training and sampling processes.
\begin{algorithm}[htb]
\caption{Autoregressive Bernoulli Spike Sampling for Posterior}
\label{alg:algorithm1}
\textbf{Input}:\\
\qquad $\bm{x}_{1:T}^E\in \{0,1\}^{C\times T}$: outputs from SNN encoder,\\
\textbf{Parameter}:\\
\qquad $f_q$: autoregressive SNN for posterior,\\
\qquad $\Theta_q$: sets of membrane potentials of $f_q$. \\
\textbf{Output}:\\
$\bm{z}_{q,1:T}\in \{0,1\}^{C\times T}$: latent spike trains.\\
$\bm{\zeta}_{q,1:T}\in \{0,1\}^{kC\times T}$: outputs of $f_q$\\
\begin{algorithmic}[1] 
\Function{PosteriorSampling}{$\bm{x}_{1:T}^E$}
\State Let $\bm{z}_{q,0}=\bm{0} \in \{0,1\}^C$
\For{$t=1$ to $t=T$}
\State $\bm{\zeta}_{q,t} = f_q (\bm{z}_{q,t-1}, \bm{x}_t^E ; \Theta_{q,t}) \in \{0,1\}^{kC}$
\For{$c=1$ to $c=C$}
\State $\bm{z}_{q,t,c} = \mathrm{random\_select}(\bm{\zeta}_{q,t}[k(c-1):kc])$
\EndFor
\EndFor
\State \textbf{return} $\bm{z}_{q,1:T}$, $\bm{\zeta}_{q,1:T}$
\EndFunction
\end{algorithmic}
\end{algorithm}

\begin{algorithm}[htb]
\caption{Autoregressive Bernoulli Spike Sampling for Prior}
\label{alg:algorithm2}
\textbf{Input}: None\\
\textbf{Parameter}:\\
\qquad $f_p$: autoregressive SNN for prior,\\
\qquad $\Theta_p$: sets of membrane potentials of $f_p$. \\
\textbf{Output}:\\
$\bm{z}_{p,1:T}\in \{0,1\}^{C\times T}$: latent spike trains.\\
$\bm{\zeta}_{p,1:T}\in \{0,1\}^{kC\times T}$: outputs of $f_p$\\
\begin{algorithmic}[1] 
\Function{PriorSampling}{\null}
\State Let $\bm{z}_{p,0}=\bm{0} \in \{0,1\}^C$
\For{$t=1$ to $t=T$}
\State $\bm{\zeta}_{p,t} = f_p (\bm{z}_{p,t-1}; \Theta_{p,t}) \in \{0,1\}^{kC}$
\For{$c=1$ to $c=C$}
\State $\bm{z}_{p,t,c} = \mathrm{random\_select}(\bm{\zeta}_{p,t}[k(c-1):kc])$
\EndFor
\EndFor
\State \textbf{return} $\bm{z}_{p,1:T}$, $\bm{\zeta}_{p,1:T}$
\EndFunction
\end{algorithmic}
\end{algorithm}

\begin{algorithm}[t!]
\caption{Overall Training and Sampling Algorithm}
\label{alg:algorithm3}
\textbf{Input}: $x$: input image\\
\textbf{Output}: $\hat{x}$: reconstructed image
\begin{algorithmic}[1] 
\Function{Training}{$x$}
\State $\bm{x}_{1:T}$ = \Call {DirectInputEncoding}{$x$}
\State $\bm{x}_{1:T}^E$ = SNNEncoder($\bm{x}_{1:T}$)

\State $\bm{z}_{q,1:T}$, $\bm{\zeta}_{q,1:T}$ = \Call{PosteriorSampling}{$\bm{x}_{1:T}^E$}
\State $\bm{z}_{p,1:T}$, $\bm{\zeta}_{p,1:T}$ = \Call{PriorSampling}{\null}
\State $\hat{\bm{x}}_{1:T}$ = SNNDecoder($\bm{z}_{q,1:T}$)
\State $\hat{x}$ = \Call{SpikeToImageDecode}{$\hat{\bm{x}}_{1:T}$}

\State calculate $\mathcal{L}$ with ($x$, $\hat{x}$, $\bm{\zeta}_{q,1:T}$, $\bm{\zeta}_{p,1:T}$) according to Eq. (25)
\State backward with $\mathcal{L}$ and update parameters

\State \textbf{return} $\hat{x}$
\EndFunction
\end{algorithmic}

\textbf{Input}: None\\
\textbf{Output}: $\hat{x}$: generated image
\begin{algorithmic}[1] 
\Function{Sampling}{\null}
\State $\bm{z}_{p,1:T}$, $\bm{\zeta}_{p,1:T}$ = \Call{PriorSampling}{\null}
\State $\hat{\bm{x}}_{1:T}$ = SNNDecoder($\bm{z}_{p,1:T}$)
\State $\hat{x}$ = \Call{SpikeToImageDecode}{$\hat{\bm{x}}_{1:T}$}

\State \textbf{return} $\hat{x}$
\EndFunction
\end{algorithmic}
\end{algorithm}

\clearpage
\onecolumn
\section{F\quad Reconstructed Images}

\begin{figure*}[ht]
 \centering
 \includegraphics[width=1\textwidth]{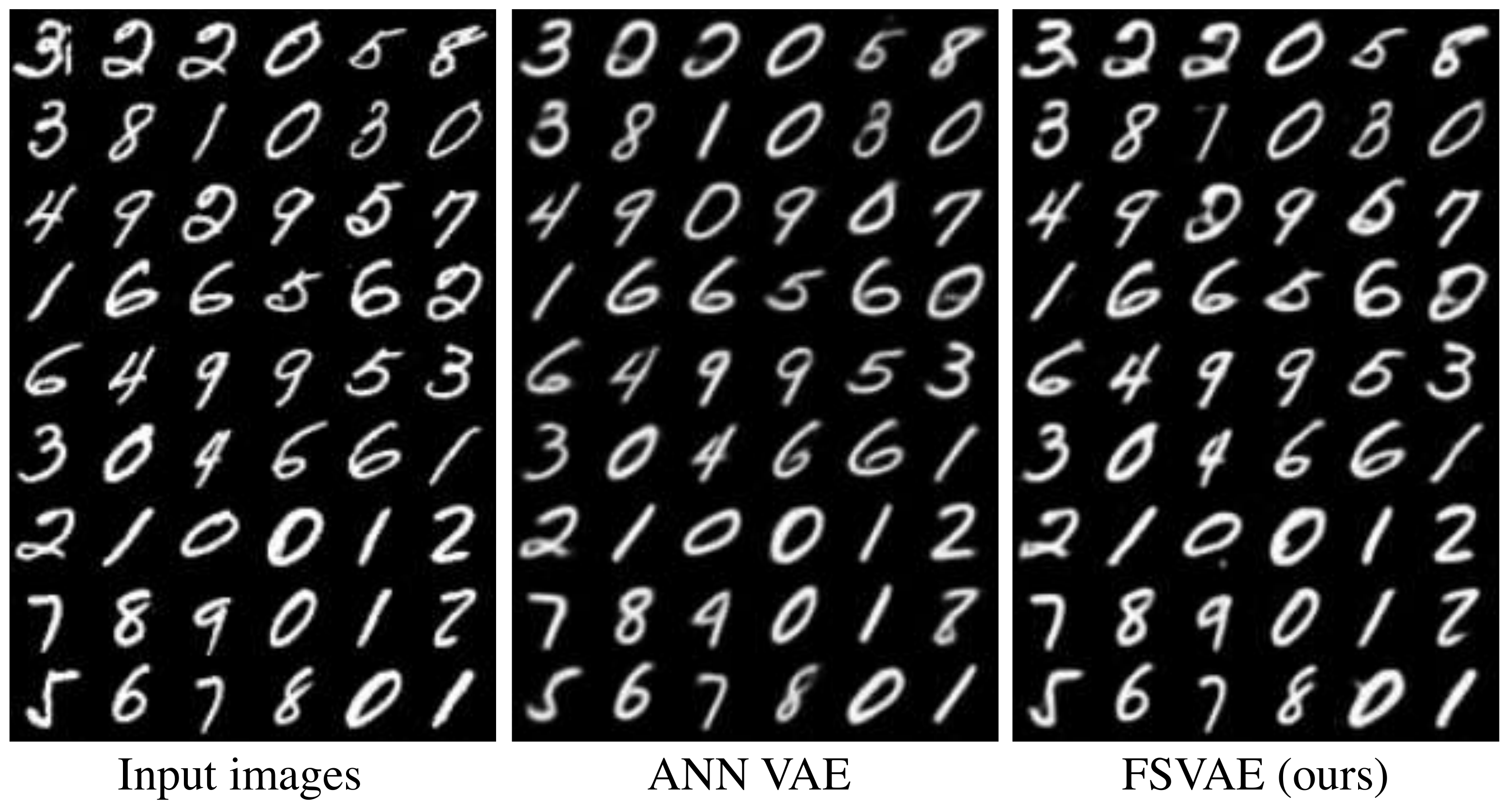}
  \caption{Reconstructed images of MNIST.}
  \label{fig:mnist_recons}
\end{figure*}

\begin{figure*}[b!]
 \centering
 \includegraphics[width=1\textwidth]{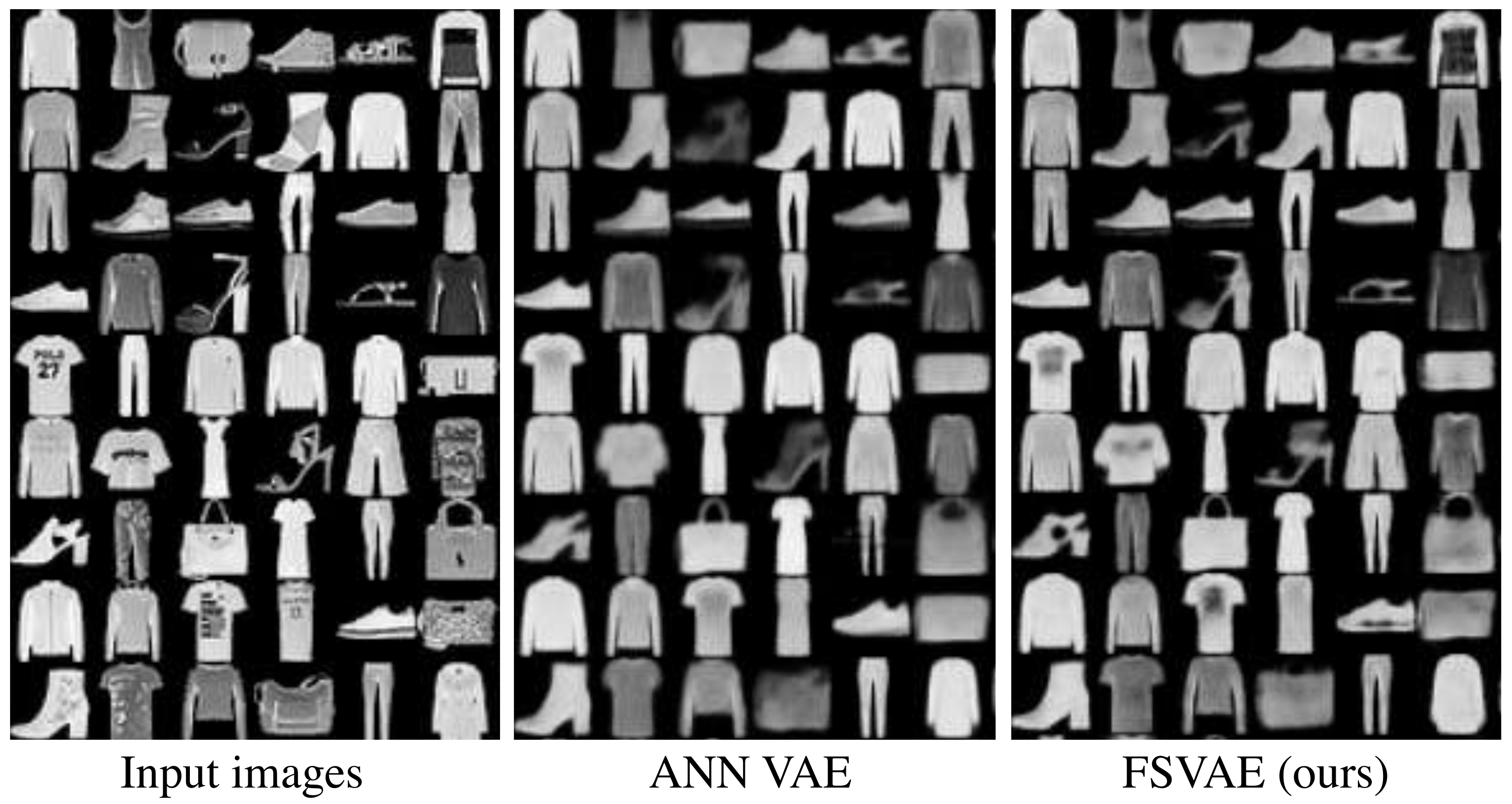}
  \caption{Reconstructed  images of Fashion MNIST.}
  \label{fig:fashion_recons}
\end{figure*}

\begin{figure*}[ht]
 \centering
 \includegraphics[width=\textwidth]{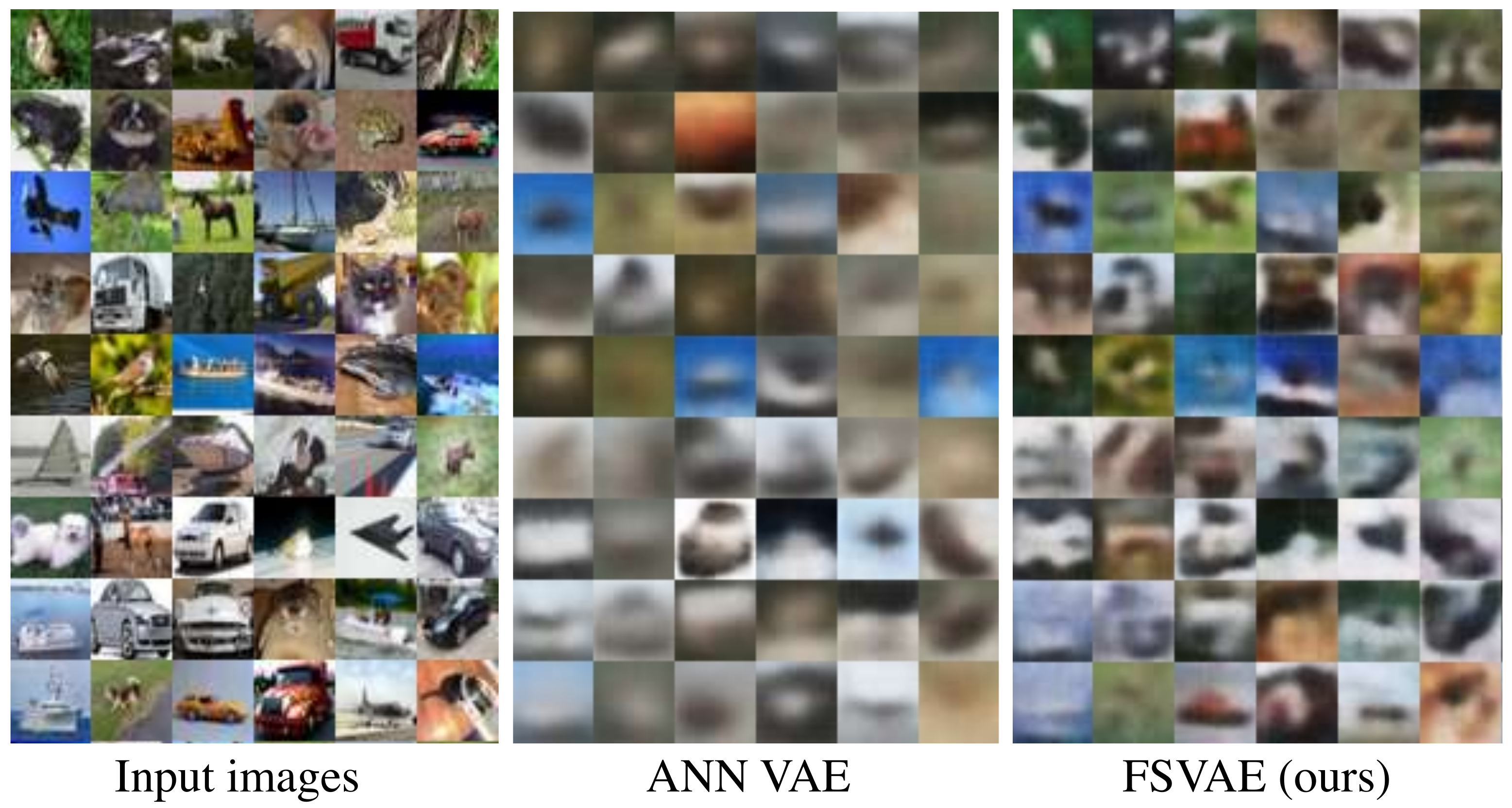}
  \caption{Reconstructed  images of CIFAR10.}
  \label{fig:cifar_recons}
\end{figure*}

\begin{figure*}[b!]
 \centering
 \includegraphics[width=\textwidth]{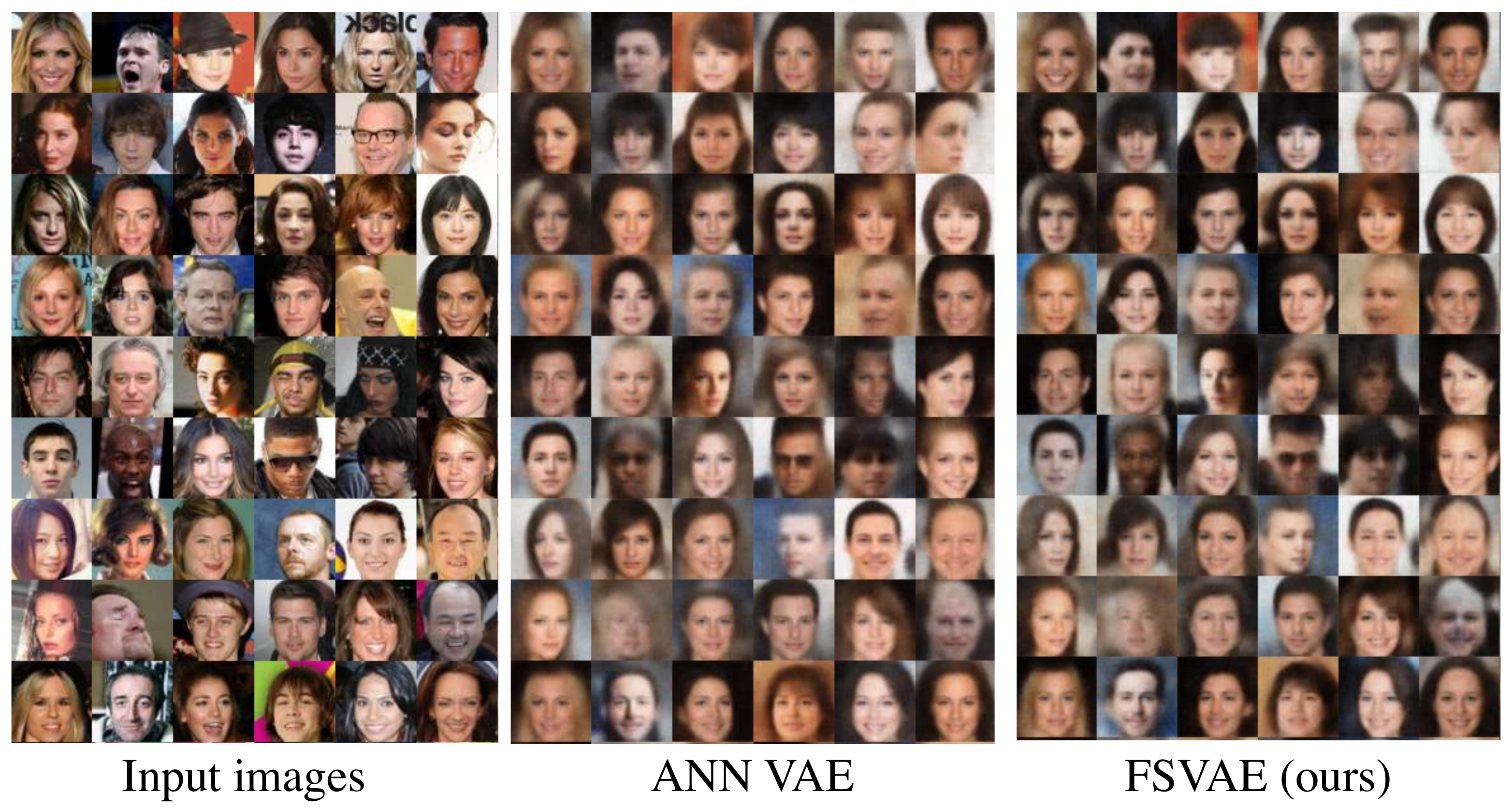}
  \caption{Reconstructed  images of CelebA.}
  \label{fig:celeb_recons}
\end{figure*}

\clearpage
\onecolumn
\section{G\quad Generated Images}

\begin{figure*}[ht]
 \centering
 \includegraphics[width=1\textwidth]{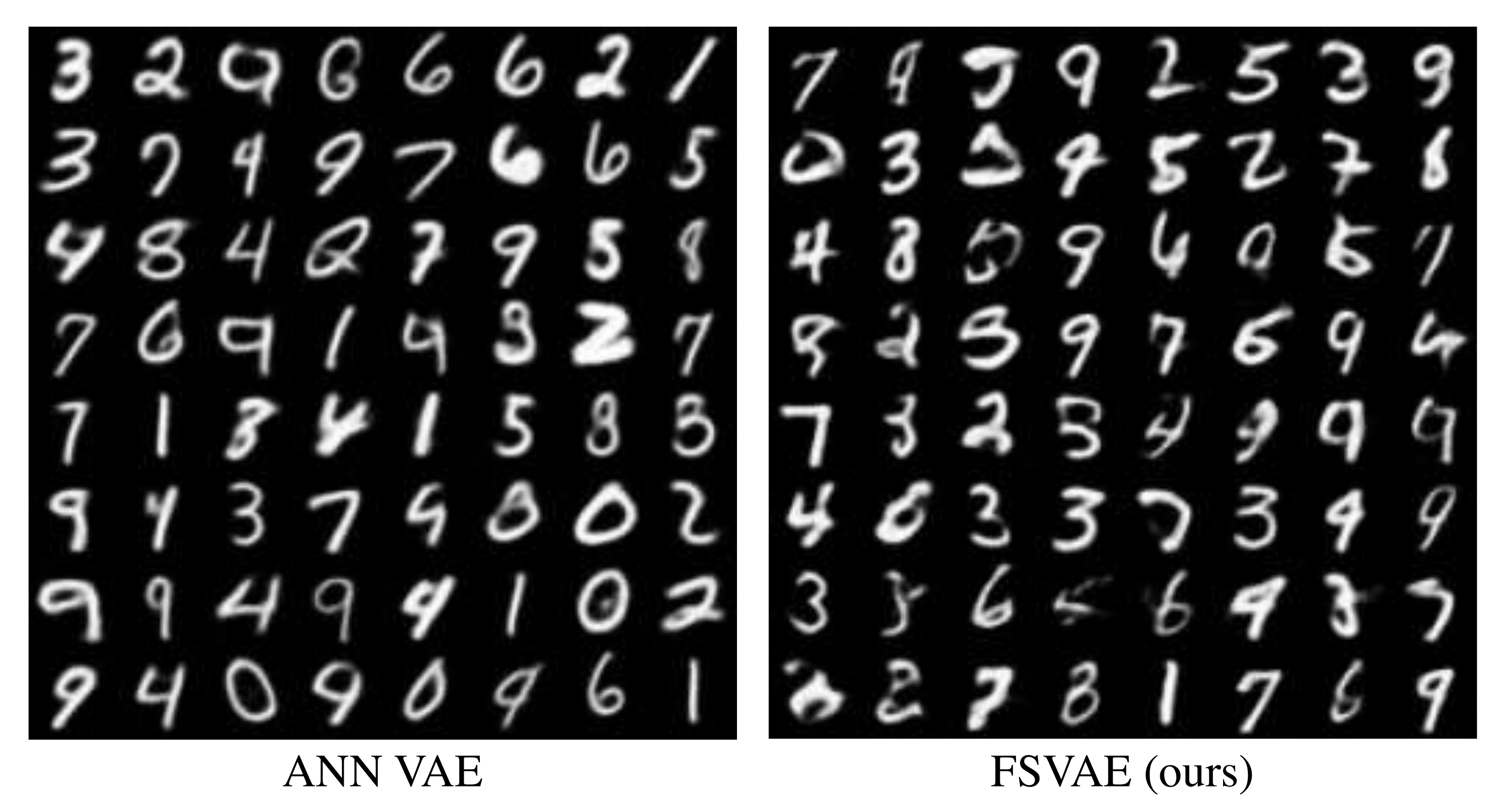}
  \caption{Generated images of MNIST.}
  \label{fig:mnist_gen}
\end{figure*}

\begin{figure*}[b!]
 \centering
 \includegraphics[width=1\textwidth]{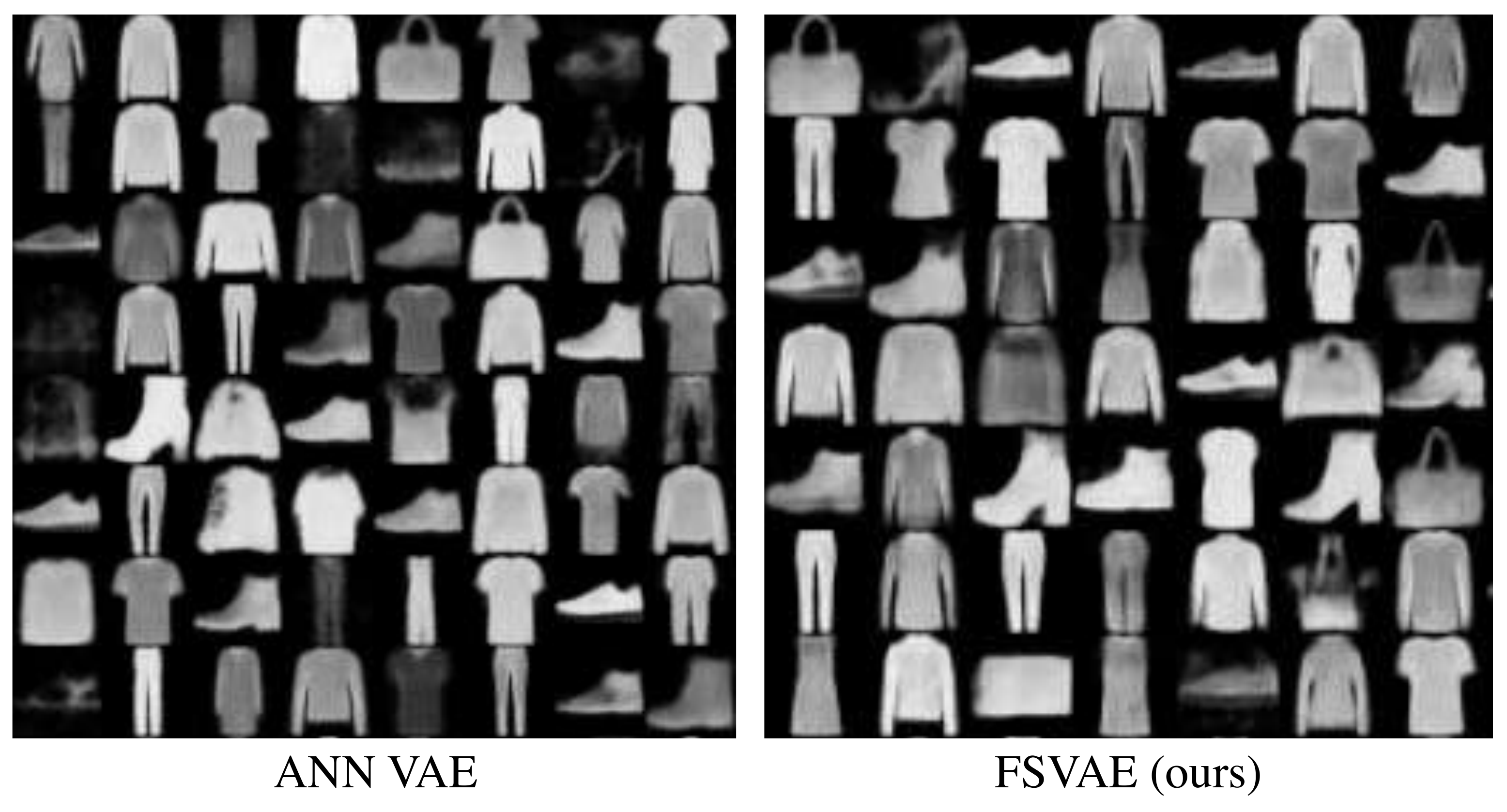}
  \caption{Generated images of Fashion MNIST.}
  \label{fig:fashion_gen}
\end{figure*}

\begin{figure*}[ht]
 \centering
 \includegraphics[width=\textwidth]{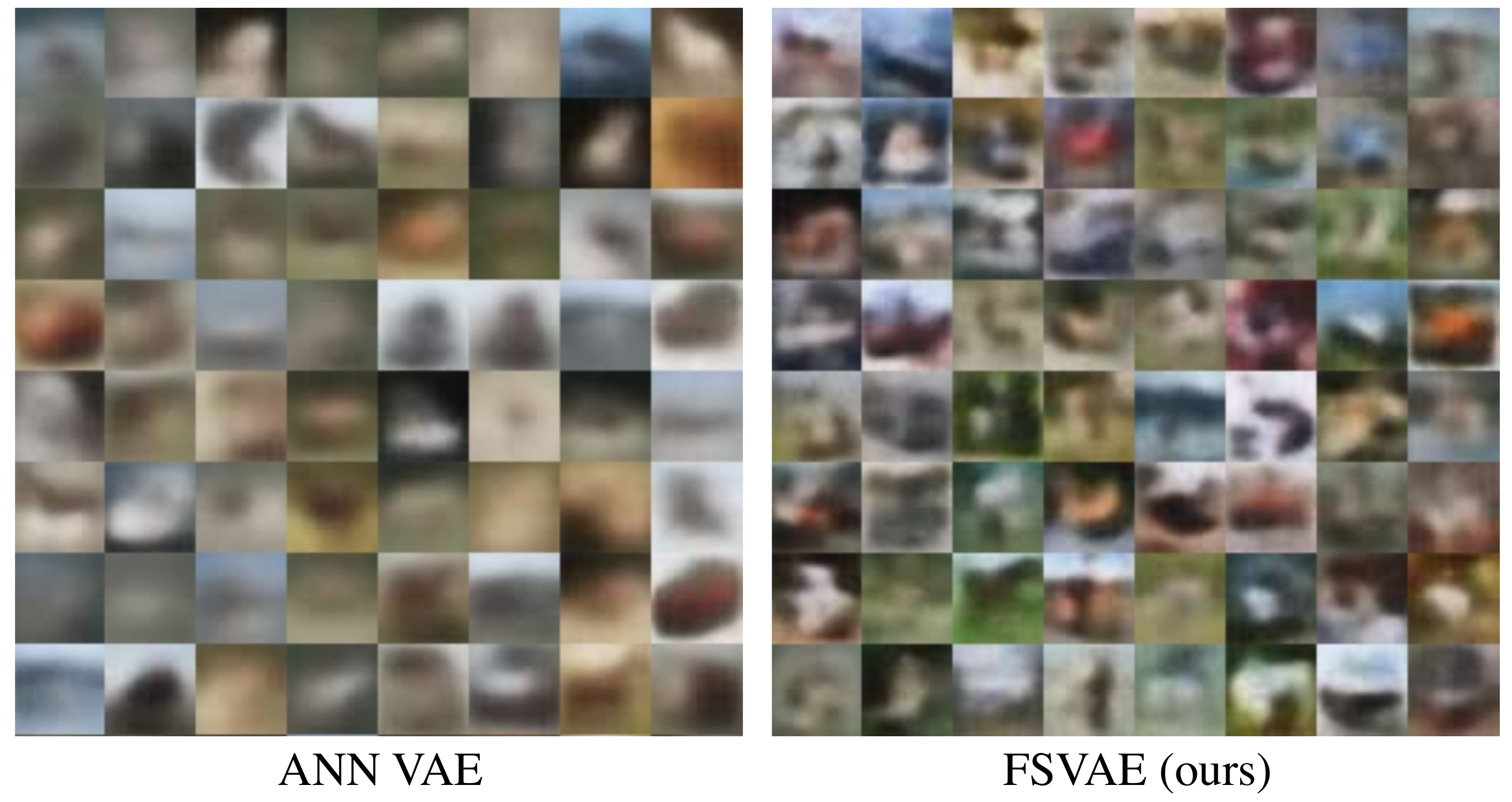}
  \caption{Generated images of CIFAR10.}
  \label{fig:cifar_gen}
\end{figure*}

\begin{figure*}[b!]
 \centering
 \includegraphics[width=\textwidth]{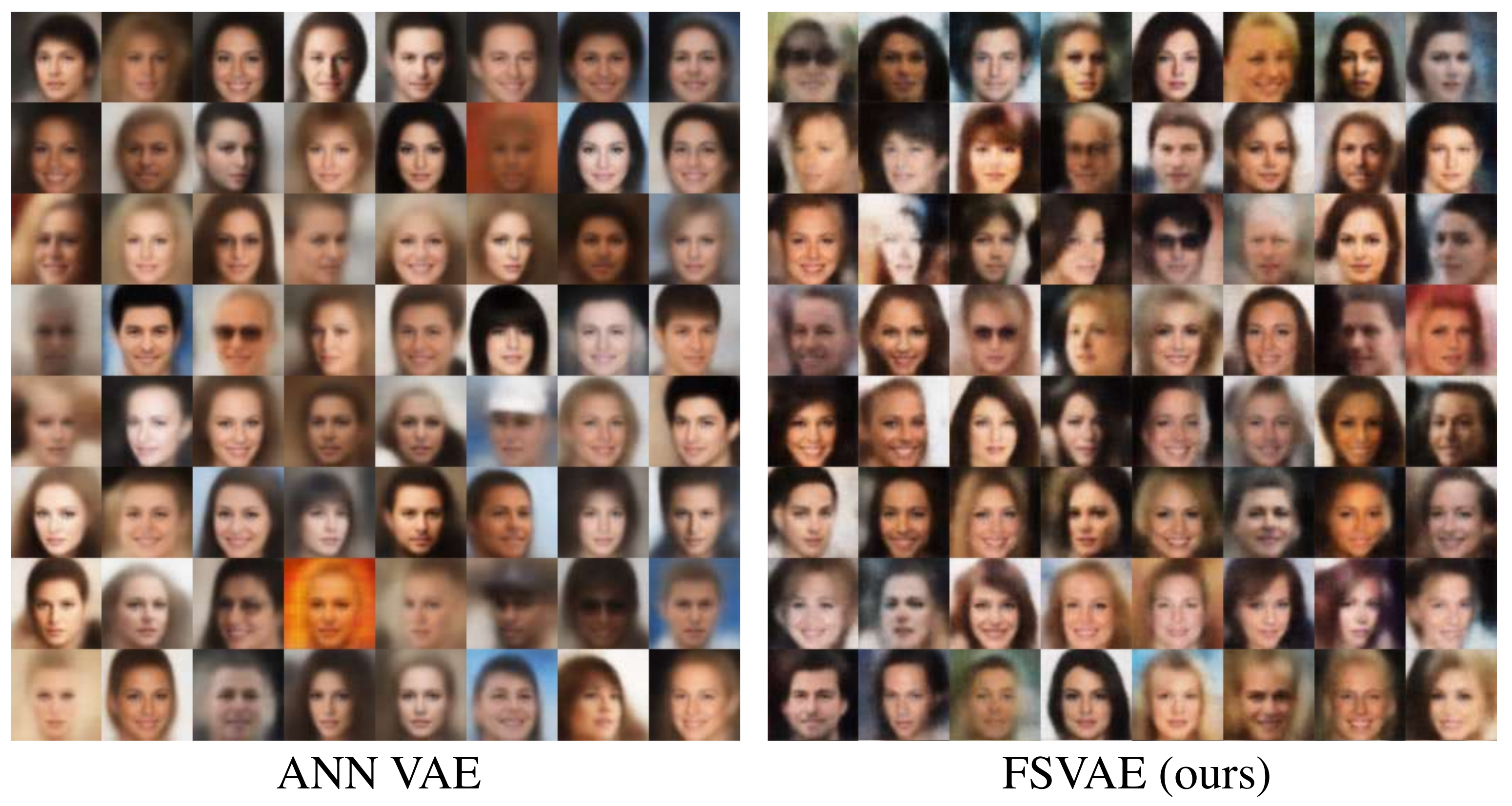}
  \caption{Generated images of CelebA.}
  \label{fig:celeb_gen}
\end{figure*}

\end{document}